\documentclass[lettersize,journal]{IEEEtran}
\usepackage{graphicx} 
\usepackage{algorithm}
\usepackage{algorithmic}
\usepackage{amsmath}
\usepackage{amssymb}
\usepackage{pifont}

\usepackage{tikz}
\usepackage{ifthen}

\DeclareRobustCommand\onedot{\futurelet\@let@token\@onedot}
\def\@onedot{\ifx\@let@token.\else.\null\fi\xspace}

\def\ie{\emph{i.e}\onedot}

\def\etal{\emph{et al}\onedot}
\makeatother

\usepackage{booktabs}
\usepackage{multirow}
\usepackage{xcolor}

\newcommand{\act}[1]{{\tt\MakeUppercase{#1}}}

\usepackage[utf8]{inputenc} 

\title{How Secure Are Large Language Models (LLMs) for Navigation in Urban Environments?}

\author{
Congcong Wen \IEEEmembership{Member, IEEE}, Jiazhao Liang, Shuaihang Yuan, Hao Huang, Geeta Chandra Raju Bethala, Yu-Shen Liu, Mengyu Wang, Anthony Tzes, Yi Fang   \\

\thanks{Congcong Wen, Jiazhao Liang, Shuaihang Yuan, Hao Huang, Geeta Chandra Raju Bethala, and Yi Fang are with Embodied AI and Robotics (AIR) Lab, New York University, New York, USA and NYUAD Center for Artificial Intelligence and Robotics, New York University Abu Dhabi, Abu Dhabi, UAE.
{\tt\small \{cw3437, jl9356, sy2366, hh1811, gb2643, yf23\}@nyu.edu}}

\thanks{Yu-Shen Liu is with the School of Software, Tsinghua University, Beijing, P. R. China. {\tt\small liuyushen@tsinghua.edu.cn}}%

\thanks{Mengyu Wang is with the Harvard AI and Robotics Lab, Harvard University, Boston, USA. {\tt\small mengyu\_wang@meei.harvard.edu}}%

\thanks{Anthony Tzes is with the NYUAD Center for Artificial Intelligence and Robotics, New York University Abu Dhabi, Abu Dhabi, UAE. {\tt\small anthony.tzes@nyu.edu}}%

\thanks{Congcong Wen, Jiazhao Liang, and Shuaihang Yuan contributed equally to this work). }
}

\begin{document}
\maketitle


\begin{abstract}
In the field of robotics and automation, navigation systems based on Large Language Models (LLMs) have recently demonstrated impressive performance. However, the security aspects of these systems have received relatively less attention. This paper pioneers the exploration of vulnerabilities in LLM-based navigation models in urban outdoor environments, a critical area given the widespread application of this technology in autonomous driving, logistics, and emergency services. Specifically, we introduce a novel Navigational Prompt Attack that manipulates LLM-based navigation models by perturbing the original navigational prompt, leading to incorrect actions. Based on the method of perturbation, our attacks are divided into two types: Navigational Prompt Insert (NPI) Attack and Navigational Prompt Swap (NPS) Attack. We conducted comprehensive experiments on an LLM-based navigation model that employs various LLMs for reasoning. Our results, derived from the Touchdown and Map2Seq street-view datasets under both few-shot learning and fine-tuning configurations, demonstrate notable performance declines across seven metrics in the face of both white-box and black-box attacks. Moreover, our attacks can be easily extended to other LLM-based navigation models with similarly effective results. These findings highlight the generalizability and transferability of the proposed attack, emphasizing the need for enhanced security in LLM-based navigation systems. As an initial countermeasure, we propose the Navigational Prompt Engineering (NPE) Defense strategy, which concentrates on navigation-relevant keywords to reduce the impact of adversarial attacks. While initial findings indicate that this strategy enhances navigational safety, there remains a critical need for the wider research community to develop stronger defense methods to effectively tackle the real-world challenges faced by these systems.
\end{abstract}



\section{Introduction}

Robot navigation technology, a pivotal component in the modern automation landscape, empowers robots to independently move and execute tasks in diverse environments. This technology has become an indispensable core element in various domains such as autonomous driving~\cite{cui2023drivellm}, warehouse logistics~\cite{kenk2019human}, household services~\cite{eirale2022human}, and emergency rescue operations~\cite{colas20133d}. With the advent of foundational models in natural language processing and computer vision, robot navigation has seen significant enhancements. Notably, large language models (LLMs)~\cite{de2023llmr} possess unique advantages in distilling vast amounts of unstructured data into actionable insights, understanding complex natural language commands, and making real-time decisions. Applying LLMs to robot navigation tasks enhances robots' reasoning and decision-making abilities, enabling more accurate processing and interpretation of real-time visual data, which is crucial for navigation in dynamic and unstructured environments.

\begin{figure}[t]
    \centering
    \includegraphics[width=1.0\linewidth]{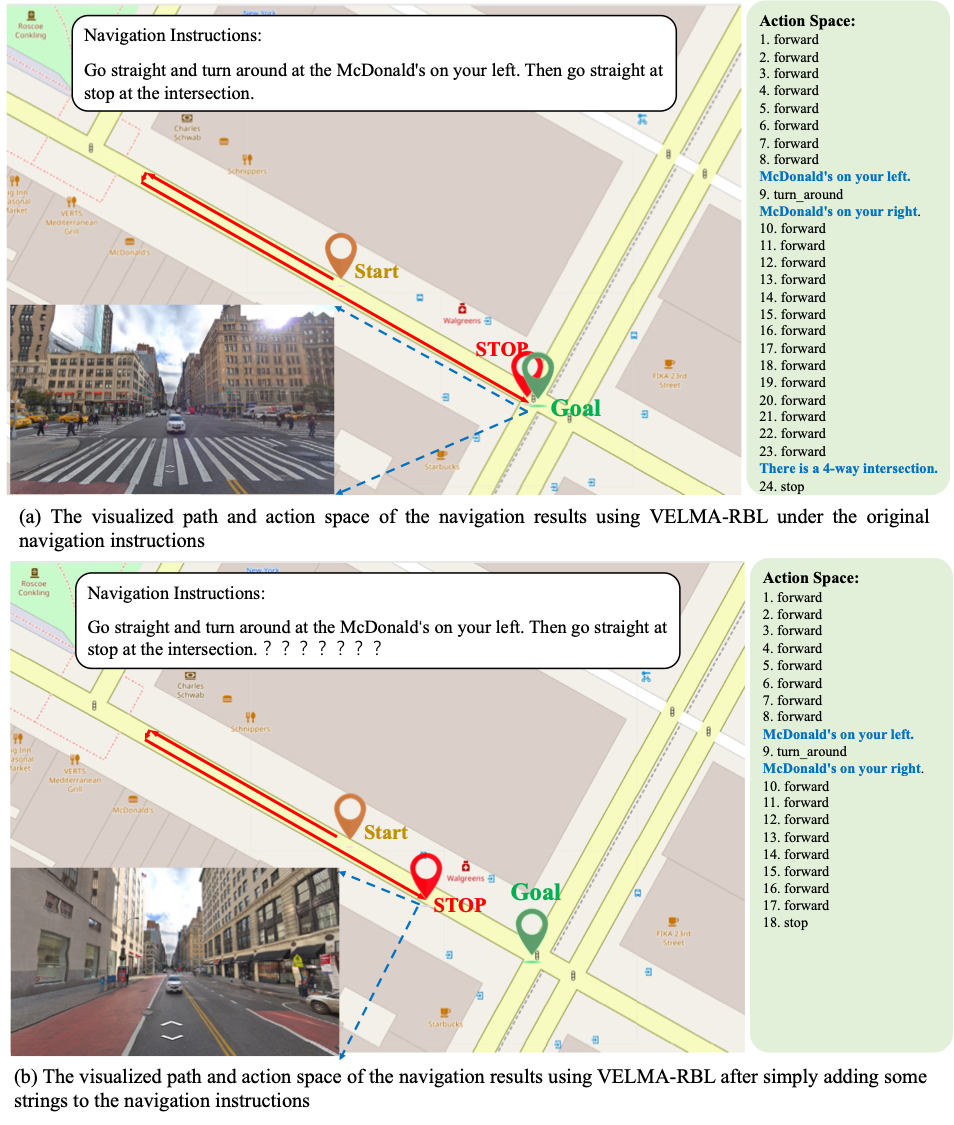}
    \caption{Comparison of navigation results of the VELMA-RBL~\cite{schumann2023velma} model before and after simple modifications to the navigation instructions.}
    \label{fig:intro}
\end{figure}

Depending on the application context, existing works utilizing LLMs for robot navigation can be categorized into indoor and outdoor scenarios. Indoor navigation tasks~\cite{yu2023l3mvn,zhou2023navgpt,yu2023co} based on LLMs typically involve directing robots to locate specific objects using textual input. In contrast, outdoor navigation tasks~\cite{shah2023lm,cui2023drivellm,schumann2023velma} often entail guiding robots through a series of instructions, including landmarks, to reach designated destinations. Although these approaches have shown promising results, most indoor navigation research relies on simulator-generated scenes or faces constraints in scale and trajectory length~\cite{schumann2023velma}. Outdoor scenarios, especially in urban environments, present unique practical challenges due to their complexity and diversity. Therefore, this paper focuses mainly on outdoor navigation in urban environments.

However, previous works have concentrated on designing LLM-based navigation systems, primarily aimed at enhancing navigation performance under normal conditions. A critical aspect that has received comparatively less attention is the inherent vulnerability of these LLM-based navigation models to potential threats. Previous studies~\cite{meng2022adversarial} have already demonstrated that most deep learning models, while robust in standard environments, can be easily deceived or disrupted when faced with carefully crafted adversarial inputs. Moreover, recent research~\cite{shi2023large} has revealed that LLMs, despite their advanced capabilities in processing and interpreting complex inputs, are exceptionally sensitive to a wide range of input variations. This sensitivity can lead to unexpected and erroneous outcomes, especially when LLMs are used as the primary decision-making framework in navigation systems. The potential for adversarial attacks to exploit these vulnerabilities and compromise the integrity of LLM-based navigation systems cannot be overlooked and needs to be thoroughly explored.

In this paper, we explore for the first time the vulnerabilities of LLM-based navigation models in outdoor urban environments, as shown in Figure \ref{fig:intro}. Inspired by previous adversarial attack strategies, we introduce a novel Navigational Prompt Attack tailored to the LLM-based navigation model. This form of attack involves perturbing the original navigational prompt by deriving gradients from the LLM model, causing the navigation model to predict incorrect actions. Based on the method of perturbation, our attacks are divided into two types: Navigational Prompt Insert (NPI) Attack and Navigational Prompt Swap (NPS) Attack. To validate the effectiveness of the proposed attacks, we conducted experiments on several variants of the VELMA model~\cite{schumann2023velma}, which utilizes diverse LLMs for reasoning, in both few-shot learning and fine-tuning experimental configurations.  The results indicated a significant decline in model effectiveness, uniformly observed across a range of attack scenarios, including both white-box and black-box methods, as evidenced by seven distinct metrics on two real-world street view datasets, Touchdown and Map2Seq. Crucially, our findings reveal the \textit{generalizability and transferability} of these attacks, where an affix generated for one LLM can effectively compromise navigation models based on different LLMs. This underscores the need for heightened security considerations in LLM-based navigation models, particularly in real-world settings like urban environments, where potential risks abound. Addressing these vulnerabilities, therefore, becomes crucial to ensure the reliability and security of LLM-based navigation systems in real-world applications. Furthermore, in response to the reduced reasoning capabilities resulting from these attacks, we have developed a preliminary countermeasure known as the Navigational Prompt Engineering (NPE) Defense strategy. This approach directs the model's attention towards keywords crucial for navigation, aiming to lessen the effects of adversarial attacks. Experimental results indicate that this defensive method can moderately offset the performance deterioration caused by such adversarial attacks. Our contribution can be summarized as follows:

\begin{itemize}
\item We are the first to investigate the safety issues of robot navigation based on Large Language Models (LLMs) in outdoor scenarios. Through extensive experiments, we demonstrate that current LLM-based navigation algorithms are highly vulnerable, highlighting an urgent need for community attention.

\item We propose a series of metrics to measure the effectiveness of attacks for robot navigation in outdoor scenarios. These metrics effectively reflect the changes in navigation performance before and after the attacks.

\item We introduce a novel Navigational Prompt Attack method, which attacks LLM-based navigation models by perturbing the original navigational prompt. This method includes the Navigational Prompt Insert (NPI) Attack and the Navigational Prompt Swap (NPS) Attack. Experimental results show that our attacks significantly degrade navigation performance across all metrics, with some metrics showing a performance drop of up to 150\%.

\item In response to our proposed attacks, we present three preliminary Navigational Prompt Engineering (NPE) defense strategies. These strategies focus on navigation-relevant keywords to mitigate the impact of adversarial attacks. Results indicate that our defense strategies can alleviate the effects of the proposed attacks to a certain extent.
\end{itemize}

\section{Related Work}

\subsection{Large Language Models for Navigation} 
Recently, large language models (LLMs) have shown robustness and effectiveness for various tasks and applications. One of the remarkable tasks that has drawn more attention is navigation. The integration of LLMs into navigation systems demonstrates a significant shift from structured indoor environments to more unpredictable outdoor settings. Previous works have already investigated how to apply LLMs to indoor navigation in simulation or real-world environments. Yu \etal~\cite{yu2023l3mvn} demonstrates the potential of LLMs to improve robotic indoor navigation, and the proposed framework is tested in both simulated and real-world environments. Co-NavGPT~\cite{yu2023co} uses LLMs for global planning and assigning exploration frontiers among robots in a multi-robot navigation problem. Zhou \etal~\cite{zhou2023navgpt} proposes NavGPT, a purely LLM-based instruction following a navigation agent, which performs a zero-shot sequential indoor action prediction for vision-and-language(VLN) navigation. Shah \etal~\cite{shah2023navigation} integrates LLms to improve indoor robotic navigation tasks by proposing a Language Frontier guide (LFC), which uses LLMs to create semantic heuristics for planning and the planning guides the robot to explore unexplored areas more effectively. Note that these works are mainly based on simulations in which either the computer-generated images are with a fixed set of displayable objects and textures, or are limited in scale and trajectory lengths. These works usually suffer sharp performance drops when testing in unseen environments~\cite{schumann2022analyzing}. The emphasis in navigation technology is increasingly shifting toward zero-shot navigation, which seeks to equip robots with the capability to explore environments without any prior knowledge. In this context, Shah et al. \cite{shah2023navigation} introduce a method called semantic guesswork, which employs a large language model (LLM) to predict promising exploration frontiers that are likely to contain target objects. Similarly, the ESC framework \cite{zhou2023esc} utilizes the commonsense reasoning capabilities of LLMs to transform semantic scenes into natural language descriptions, assisting the robot in determining the next exploration point. Additionally, Song et al. \cite{song2023llm} apply an LLM as a zero-shot/few-shot planner that generates a series of actions for the robot, which are subsequently refined through a closed-loop feedback mechanism. In addition to indoor navigation, outdoor navigation is more complicated due to the realistic and unpredictable elements of these environments. In outdoor navigation, landmarks are first extracted, and then new instructions with descriptions of related landmarks are provided. Then, LLMs are responsible for predicting the agent's next step. Our work focuses mainly on outdoor navigation. Schumann \etal~\cite{schumann2023velma} demonstrates the applicability of LLMs in real-world scenarios. The landmarks are first extracted using CLIP, and then the fine-tuned models based on Llama~\cite{touvron2023llama} are required to make decisions when a robot comes to intersections. Shah \etal~\cite{shah2023lm} proposes LM-Nav, where GPT-3~\cite{brown2020language} is used for outdoor navigation. Cui~\etal~\cite{cui2023drivellm} also proposes a decision-making framework that combines LLMs with autonomous driving systems, enabling commonsense reasoning and learning from mistakes through a cyber-physical feedback system.

\subsection{Security of Large Language Models}
With the expanding scope of LLM applications, their security has become a critical area of focus. Yao \etal~\cite{yao2023survey} categorizes LLM vulnerabilities into two categories: \textit{Inherent Vulnerabilities of the AI Model} and \textit{Vulnerabilities not Inherent to the AI Model}. The latter is unrelated to the LLM itself and usually requires a lot of extensive engineering tests to find such vulnerabilities. In contrast, inherent vulnerabilities can typically be detected through adversarial attacks. Wang \etal~\cite{wang2021towards} conducts an extensive survey on adversarial techniques to generate adversarial texts and designs the corresponding defense methods. Papernot \etal~\cite{papernot2017practical} introduces an adaptive black-box attack, and this method is further refined by Mahmood \etal~\cite{mahmood2021besting} which enhanced the attack and demonstrated $\geq30\%$ more effective than~\cite{papernot2017practical}. Zeng \etal~\cite{zeng2024johnny} proposes interpretable adversarial prompts to challenge AI safety for LLMs. A key difficulty in launching adversarial attacks on LLMs lies in the intrinsic discreteness of text, as opposed to the continuous nature of image input.

This distinction makes it difficult to employ gradient-based optimization methods to effectively craft adversarial attacks. To solve such a problem, two solutions are suggested. One is embedding-based optimization, also known as soft prompting~\cite{lester2021power}, projects discrete tokens in some continuous embedding space. Unfortunately, the challenge is that the process is not reversible and LLMs usually do not allow users to provide continuous embeddings. The Prompts Made Easy (PEZ) algorithm~\cite{wen2023hard} solves these problems by using a quantized optimization that can project the result back into the hard prompt space. Another is one-hot encoding, which treats the one-hot vector as if it were a continuous quantity, as first proposed in HotFlip~\cite{ebrahimi2017hotflip}. However, the gradients at the one-hot encoding level may not provide an accurate representation of the function's behavior after an entire token is replaced. To solve the above problems, AutoPrompt~\cite{shin2020autoprompt} proposes a method based on gradient-guided search to automatically design prompts, which maximize the likelihood of prompting by constant iterative updating, and keep the prompt with the highest probability in the next step. Recently, ARCA~\cite{jones2023automatically} further improved this by evaluating not only the approximate one hot gradients at the original one-hot encoding of the current token, but also at several potential token swaps. More recently, Zou \cite{zou2023universal} proposed a method to jailbreak LLM using automatically generated adversarial suffixes.

In addition to the adversarial attack method, it is important to design techniques to prevent the LLM from being fooled. Among the existing method, preprocessing input prompts using methods like perplexity filters \cite{jain2023baseline}, and paraphrasing \cite{jain2023baseline} have shown some effectiveness. Concurrent research also investigates heuristic detection-based methods, which have demonstrated strong performance in identifying adversarial inputs \cite{alon2023detecting}. Another innovative approach involves the use of safety filters on substrings of input prompts, offering certifiable robustness guarantees \cite{kumar2023certifying}. Although this method shows promise, its complexity scales with prompt length, posing a challenge for practical implementation. Robey~\etal \cite{robey2023smoothllm} contributes to this body of work by developing a defense mechanism that perturbs input prompts at the character level, aggregating responses to detect and mitigate adversarial attacks.

\subsection{Prompt Engineering Technique}

With the release of the GPT series, the utilization of large language models as alternatives to supervised learning has acquired significant significance, as stated in~\cite{radford2019language}. Parallel to the rapid evolution of substantial large language models~\cite{brown2020language,ouyang2022training}, the engineering of prompts~\cite{gao2021making,schick2021s,weng2023towards} emerged as a prospective framework for model interactions~\cite{lester2021power}. By allowing users to focus on the design and refinement of prompt inputs, this framework makes it possible to improve the effectiveness of the pre-trained model in different applications. This is accomplished by directly utilizing the intrinsic knowledge stored in the pre-trained model, thereby avoiding any additional training phase. The application of prompt engineering has been widely studied~\cite{gao2021making,le2021many,schick2021s,wei2022chain}. This research has been conducted in both zero-shot and few-shot scenarios. There are three different types of prompting approach: the first is to use human-comprehensible prompts in large-scale models, and the second is to tune prompt, which involves learning a continuous prompt representation~\cite{lester2021power,li2021prefix}. The third type of prompting approach is auto-prompting~\cite{shin2020autoprompt}, which involves the autonomous generation of prompts from the ground up. Both of the latter two approaches require a phase of training. In the first type of prompting strategy, the user is often responsible for the work of developing efficient prompts to extract certain information from a model. This task frequently involves a trailing and error process. For the second and third prompt approaches, PromptSource~\cite{bach2022promptsource} enhances the capabilities of the templating syntax utilized in the formulation of generic prompts. More specifically, it provides a platform for the communal development, evaluation, and discovery of new prompts. Using the generative features of the T5 model \cite{raffel2020exploring}, Gao \etal~\cite{gao2021making} constructs prompt templates and accelerates the exhaustive search process to choose the label words. PromptIDE~\cite{strobelt2022interactive} offers users the opportunity to utilize interactive visual tools that assist them in evaluating the success of prompts within a small dataset and improve the cyclical functioning of the prompts. To improve the transparency and manageability of large language models, PromptChainer~\cite{wu2022promptchainer} enables users to constructively compose sequences of prompts that are suitable for specific subtasks. This is particularly useful for complex tasks that need multi-phased operations. There are more examples that can be found in a recent review by Liu \etal~\cite{liu2023pre}. Recent research also demonstrates that LLMs can also work as decent zero-shot reasoners~\cite{kojima2022large} by being prompted using a chain of prompts to endow LLMs with reasoning ability~\cite{wang2023plan,kong2023better}.

\section{Vulnerabilities in LLMs for Navigation}

\begin{figure*}[t]
    \centering
    \includegraphics[width=0.95\linewidth]{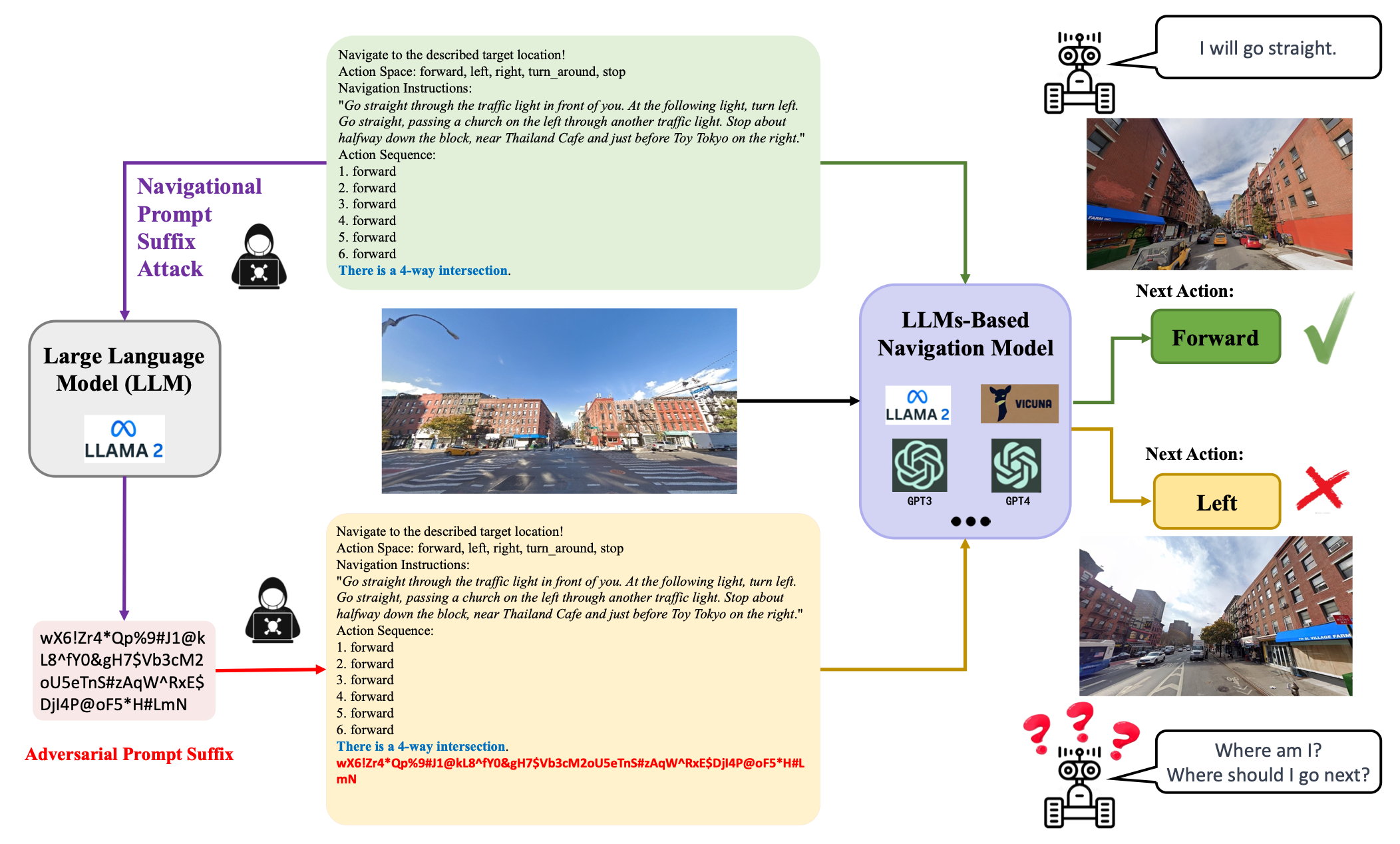}
    \caption{Illustration of the Navigational Prompt Affix Attack. This approach involves attacking a specific LLM to identify an Adversarial Prompt Affix. When this affix is directly appended to a navigation prompt, it causes navigation systems based on that LLM, or other LLM models, to predict incorrect actions.}
    \label{fig:pipe}
\end{figure*}

\subsection{LLMs-based Navigation} \label{sec:llm_nav}

\subsubsection{Problem Statement} 
The primary goal of the navigation task is for the robot to determine the target location \( G \) from an initial state \( s_0 \), utilizing a set of navigation instructions denoted as \( n = (w_1, w_2, \ldots, w_N) \). In this undertaking, the robot must calculate a route that passes through a series of waypoints \( (\hat{v}_1, \hat{v}_2, \ldots, \hat{v}_T) \). The robot's state, signified as \( s = (v, \alpha) \), is characterized by its current position \( v \) among the network of nodes \( V \), and its orientation \( \alpha \). From the starting state \( s_0 \), the robot deduces the most likely next position by interpreting the navigation instructions \( w_i \) alongside the current observations \( o_i \), which encompass a panoramic view and the count of outward paths from its current node. The likelihood of the robot moving to the next position \( v_{t+1} \) is modeled by the following equation:
\begin{equation}
    v_{t+1} = \arg \max_{\bar{v} \in \bar{V}} P_{LLM}(\bar{v} | w_t, o_t)
    \label{eq:position_pred}
\end{equation}
where \( v_{t+1} \) represents the predicted next waypoint the robot aims to reach, \( \bar{V} \) is the set of all possible waypoints the robot can move to, \( w_t \) are the navigation instructions at time \( t \), and \( o_t \) are the observations at time \( t \). The function \( P_{LLM} \) provides the probability of reaching any given waypoint \( \bar{v} \) based on the current instructions and observations. The robot selects the waypoint with the highest probability as its next position.

Upon executing a move to the new position, the environment updates accordingly and transitions the robot into the subsequent state. This cycle of movement and transition continues iteratively. The robot progresses through this loop, advancing toward the designated goal, until it reaches a state where it predicts \(\act{stop} \), indicating that it has presumably arrived at the destination. The navigation task is considered successfully concluded when the robot's calculated position coincides with the target location \( G \), satisfying the navigational objective.

\subsubsection{Target Model \textbf{VELMA}} 

Translating this general framework into practical applications, we turn our attention to a specific example in the realm of LLM-based navigation. After a thorough analysis of existing navigation methods based on LLMs, we have selected {VELMA} as our target model due to its experiment settings that closely align with real-world environments. Unlike previous studies that were mainly focused on simulation environments or confined indoor spaces, VELMA is dedicated to exploring navigation tasks in actual city streetscapes, utilizing datasets such as {Touchdown} and {Map2seq}. This model demonstrates the efficacy of LLMs in long-distance navigation across real urban road networks and complex street scenarios. Therefore, we believe that hastily applying such navigation systems to the real world without a comprehensive analysis of their safety could lead to significant risks, particularly when taking into account the inherent vulnerabilities of deep learning models. Since LLMs are based on these neural network technologies, their unvetted deployment in real-world scenarios might trigger a series of societal problems and safety hazards.

As we introduced before, for the current navigation work based on LLMs, it essentially involves predicting the probability \( p(v_{i+1} | w_i, o_i) \) of the next position \( v_{i+1} \) based on the current navigation instruction \( w_i \) and observation \( o_i \). Specifically for VELMA, it has shifted the focus from predicting positions to predicting actions, where the action \( a \) is within the action space \( A \), defined as $A =\{\act{forward},\act{left},\act{right},\act{turn\_around},\act{stop}\}$. The formula for predicting the next action is expressed as:
\begin{equation}
    a_{t+1} = \arg \max_{\bar{a} \in A} P_{LLM}(\bar{a} | x_t),
    \label{eq:action_pred2}
\end{equation}
where $x_t$ represents the complete text prompt used as the input for the LLM at the current time \( t \). It is constructed by concatenating the task description \( d \), the navigation instructions text \( n \), followed by a sequence of observations \( o_i \), timestamps \( i \), and actions \( a_i \) for each timestep up to \( t-1 \), and finally appending the current observation \( o_t \) and the current timestamp \( t \). The formulation is as follows:
\begin{equation}
    x_t = [d, n] \oplus \bigoplus_{i=1}^{t-1} [o_i, i, a_i] \oplus [o_t, t],
    \label{eq:prompt3}
\end{equation}
In the VELMA model, the first step involves the Landmark Extractor Module, which leverages LLM technology to derive a sequence of landmarks $(l_1, l_2, \ldots, l_L)$ from the given navigation instructions $n$. This is followed by the utilization of the Landmark Scorer Module, integrating CLIP, to evaluate whether each landmark $l_i$ is visible in the corresponding observation $o_i$. The next phase involves the Verbalizer Module, a template-based component, which produces textual descriptions of the environmental observations. These descriptions are subsequently integrated into the comprehensive prompt $x_t$. The model then proceeds to predict the subsequent action \(a_{t+1}\) by using the prompt \(x_t\) as input for the LLM, as delineated in Eq.~\ref{eq:action_pred2}. The navigation task is considered complete when the model predicts the action \(\act{stop}\), indicating the robot considers itself to have reached the goal destination $g$.

\subsection{Navigational Prompt Attack}\label{sec:llm_attack}
As discussed in the previous section, LLMs are commonly utilized to predict the next location or action in navigation tasks, based on current instructions and observations. It is vital to ensure that these models produce reasonable and reliable outcomes. Nevertheless, the security and reliability aspects of existing LLMs, especially in the context of real-world navigation tasks, are not sufficiently explored.

In the realm of deep learning, particularly in image recognition, adversarial attacks often revolve around finding minute input data variations that can mislead the model. These variations are identified through the model's gradient. Attackers calculate the gradient of the model's output relative to the input data to craft subtle, nearly undetectable perturbations. When applied to the original input, these perturbations can cause the model to make erroneous predictions or classifications. Drawing inspiration from this methodology and related research, we have introduced two adversarial attack strategies for the LLM reasoning module used in robotic navigation tasks: \textbf{Navigational Prompt Insert (NPI) Attack} and \textbf{Navigational Prompt Swap (NPS) Attack}. These involve modifying the original prompt through insertion or swapping operations, respectively, to mislead the robot, causing it to generate incorrect location and action predictions. Our experiments show that these methods of perturbing the prompt are not only effective against specific target LLMs but also against other LLMs. This finding underscores the general applicability of such attacks. More importantly, it highlights the vulnerability of current LLMs in navigation scenarios, emphasizing the urgent need for the research community to focus more on the security aspects of deploying LLMs in real-world environments.

Specifically, the core of the navigational prompt attack is to minimize the likelihood of the LLM making an accurate prediction for the next action by perturbating the original prompt. To quantify this, we define a loss function \( \mathcal{L} \) as follows:
\begin{equation}
    \mathcal{L}(x_t, a_t) = - \log P_{LLM}(a_t | x_t)
\end{equation}
This loss function measures the negative logarithm of the probability that the LLM will correctly predict the next action \( a_t \) given the prompt \( x_t \). The objective of navigational prompt  attack is then formalized by the following optimization problem:
\begin{equation} 
    \min \mathcal{L}(\hat{x}_t, a_t),
    \label{eq:obj_attack}
\end{equation}
\noindent where \( \hat{x}_t \) represents the perturbated navigational prompt.  Depending on the type of perturbation, adversarial navigation prompts can be categorized into insertion and swapping methods. By optimizing this loss, two types of navigational attacks effectively decrease the LLM’s likelihood of making accurate predictions in navigation scenarios, thereby exposing the model’s vulnerabilities to such adversarial interventions.

\subsubsection{\textbf{Navigational Prompt Insert Attack (NPI) Attack}} NPI Attack is a strategy designed to append a carefully crafted affix to the original navigation prompt \( x_t \). The objective of the NPI Attack can be formalized by the following optimization problem:
\begin{equation}
    \min_{\bar{x} \in X} \mathcal{L}(x_t \oplus \bar{x}, a_t),
    \label{eq:obj_attack}
\end{equation}
\noindent where \( X \) represents the set of all possible adversarial affixes, \( \oplus \) denotes the operation of concatenating the affix to the prompt, and \( \bar{x} \) is the generated adversarial affix. 

Unlike attacks on continuous inputs as seen in image processing, the input in Eq.~\ref{eq:obj_attack} consists of discrete text tokens. To execute a discrete attack, previous works typically employed a method of swapping the token that maximally decreased the loss. Building on similar ideas from these works, we first generate initial prompt $\widetilde{x}$ by adding a set of affix tokens and then identify several possible token substitutions of affix tokens according to the $k$ largest negative gradients with respect to one-hot token indicators. The formulation is as:
\begin{equation}
\mathcal{X}_i := \mbox{Top-}k(-\nabla_{e_{\widetilde{x}_i}} \mathcal{L}(\widetilde{x})), i \in \mathcal{S}, \\
\label{eq:topk}
\end{equation}
\noindent where $\mathcal{S}$ denotes the indices of the adversarial affix tokens. By iterating over the set $\mathcal{S}$, we compile an initial set $\mathcal{X}$ of potential token substitutions. From this set $\mathcal{X}$, we then randomly select a subset $\mathcal{X}_B$ of tokens as candidates for replacement. The final step entails selecting the optimal replacement token from the subset $\mathcal{X}_B$ that minimizes the loss function $\mathcal{L}$. This process is expressed as the following formula:
\begin{equation}
\bar{x} = \underset{\bar{x} \in \mathcal{X}_B}{\mathrm{argmin}}\, \mathcal{L}(x_t \oplus \bar{x}, a_t),
\end{equation}
\noindent where $\bar{x}$ represents the token that, when replaced in the prompt, yields the smallest value of the loss function $\mathcal{L}$, thereby identifying the most advantageous adversarial token replacement.

\subsubsection{\textbf{Navigational Prompt Swap Attack (NPS) Attack}} Unlike the NPI Attack, which appends a affix directly to the original prompt without altering it, we have devised an alternative strategy that involves replacing words in the original prompt. Specifically, we aim to replace a word in the original prompt to decrease the likelihood that the navigation model will output the correct action. This optimization process can be described as follows:
\begin{equation} 
    \min_{\bar{x} \in X} \mathcal{L}(x_t \otimes \bar{x}, a_t),
    \label{eq:obj_attack}
\end{equation}
where \( X \)  represents the set of all possible tokens in the vocabulary that can be swapped, \( \otimes \) symbolizes the operation of swapping tokens within the prompt, and \( \bar{x} \) is the token selected from the dictionary to be swapped with a word in the original prompt.

In order to minimize the perturbation to the original prompt, we replace only one word in the original prompt. To ensure that the replacement is a more meaningful word, specifically one that contains more critical navigational information, we first calculate the importance of each word. To ensure that our attack method is broadly applicable, that is, effective for both black-box and white-box attacks, we use the gradient response of the navigation model as a measure of word importance. This means we select the word with the greatest gradient response to the navigation model as the word to be swapped. Let \( x_t \) represent the original navigation prompt consisting of words \( w_1, w_2, \dots, w_n \). Our goal is to identify and replace one word in \( x_t \) to minimize the accuracy of the navigation model's output. The process can be summarized through the following steps:
\begin{enumerate} 
    \item \textbf{Calculate Word Importance:}
    We compute the importance of each word \( w_i \) in the prompt based on the gradient response with respect to the model's output. This can be represented as:
    \[
    \text{Importance}(w_i) = \left\lvert \nabla_{w_i} \mathcal{L}(x_t, a_t) \right\rvert
    \]
    Here, \( \mathcal{L} \) is the loss function of the navigation model, \( a_t \) is the desired action or output from the model, and \( \nabla_{w_i} \mathcal{L} \) represents the gradient of the loss with respect to word \( w_i \).

    \item \textbf{Select the Word to Swap:}
    We identify the word \( w_i \) with the highest gradient magnitude as the word to replace. This is because a higher gradient magnitude indicates a higher sensitivity of the output to changes in this word:
    \[
    w_{\text{swap}} = \underset{w_i}{\mathrm{argmax}}\, \text{Importance}(w_i)
    \]

    \item \textbf{Word Replacement:}
    After identifying \( w_{\text{swap}} \), the original prompt \( x_t \) can be represented as: $ x_t = x_{t, \text{b}} \oplus w_{\text{swap}} \oplus x_{t, \text{a}} $, where \( x_{t, \text{b}} \) represents the sequence of words in \( x_t \) before \( w_{\text{swap}} \), and \( x_{t, \text{a}} \) represents the sequence of words in \( x_t \) after \( w_{\text{swap}} \). We then select a swap token \( \bar{x} \) from a dictionary \( X \) using an equation referred to as Eq.~\ref{eq:topk} to swap \( w_{\text{swap}} \). The swap process can be described using the optimization formula where \( \bar{x} \) minimizes the loss function:
    \[
    \bar{x} = \underset{\bar{x} \in X}{\mathrm{argmin}}\, \mathcal{L}(x_{t, \text{b}} \oplus ( w_{\text{swap}} \otimes \bar{x} ) \oplus x_{t, \text{a}}, a_t)
    \]
\end{enumerate}

\subsection{Experiment for NPS Attack in LLMs-based Navigation}\label{sec:exp_res}

\subsubsection{Datasets} 
\paragraph{\textbf{Touchdown}} The Touchdown dataset is proposed by \cite{chen2019touchdown}. This dataset is created based on the Google Street View of New York City. The touchdown dataset contains 9,326 examples of English instructions and spatial descriptions paired with demonstrations. Each position includes a 360 degree RGB panorama and all the panoramas are connected in a graph-like structure with undirected edges connecting neighboring panoramas. The environment includes 29,641 panoramas and 61,319 edges in total. The process is split into four crowdsourcing tasks: instruction writing, target propagation to panoramas, validation, and segmentation. Here, the two main tasks are writing and following. During the writing stage, a leader follows a prescribed route and hides Touchdown Bear at the end while writing navigation instructions that describe the path. The following stage involves following the written navigation instructions from the same starting position to find the Touchdown but only showing panoramas without the map.

\paragraph{\textbf{Map2seq}} The Map2Seq Dataset is originally proposed by \cite{schumann2020generating} based on OpenStreetMap(OSM). OSM is a set of geodata that provides detailed street layouts and annotations such as amenities, infrastructure, or land use. Map2Seq Dataset exported an OSM extract of Manhattan in 2017 and discretized the street layout by creating a node every ten meters along the roads. Routes are then sampled with lengths between 35 and 45 nodes. Afterwards, they used Amazon Mechanical Turk(AMT) to acquire human annotators. The first annotator is shown a specific route mentioned above and asked to write navigation instructions, including visual landmarks, without seeing the true panorama images. Then, a different annotator validates navigation instructions by using them to follow the route in the panorama environment without the map. Eventually, the dataset contains 7,672 navigation instructions.

\begin{table*}[h] 
\centering
\caption{Results for the VELMA-FT and VELMA-RBL models before and after the \textbf{Navigational Prompt Insert (NPI) Attack} on the Touchdown and Map2seq datasets in the unseen scenario. $\dagger$ represents the results of attacks on prompts describing intersections, while $\ddagger$ indicates the results of attacks on prompts describing landmarks. $\circ$ indicates white-box attacks, and $\bullet$ indicates black-box attacks.}
\resizebox{1.0\linewidth}{!}{
\begin{tabular}{lccccccc}
\toprule
&\multicolumn{7}{c}{\textbf{Touchdown}} \\ 
\cmidrule{2-8} 
    \textbf{LLM Models} & \textbf{SPD\textdownarrow} & \textbf{KPA\textuparrow} & \textbf{TC\textuparrow} & \textbf{FKPE\textdownarrow} & \textbf{DI\textdownarrow} & \textbf{PL\textdownarrow} & \textbf{TC-1\textuparrow}\\
\toprule
\multicolumn{8}{c}{2-Shot In-Context Learning} \\
\midrule
VELMA-GPT3 &26.0 &39.6 &6.2 &289 &65 &32 &10.1\\
$\bullet$ VELMA-GPT3 $\dagger$ &27.5\scriptsize{(\textcolor{blue}{\textuparrow 5.78\%})}  &24.8\scriptsize{(\textcolor{red}{\textdownarrow 37.37\%})} &3.0\scriptsize{(\textcolor{red}{\textdownarrow 51.61\%})} &400\scriptsize{(\textcolor{blue}{\textuparrow 38.41\%})} &72\scriptsize{(\textcolor{blue}{\textuparrow 10.77\%})} &45\scriptsize{(\textcolor{blue}{\textuparrow 40.63\%})} &6.4\scriptsize{(\textcolor{red}{\textdownarrow 36.63\%})}\\
$\bullet$ VELMA-GPT3 $\ddagger$ &26.6\scriptsize{(\textcolor{blue}{\textuparrow 2.31\%})} &25.3\scriptsize{(\textcolor{red}{\textdownarrow 36.11\%})} &4.5\scriptsize{(\textcolor{red}{\textdownarrow 27.42\%})} &370{(\textcolor{blue}{\textuparrow 28.03\%})} &68\scriptsize{(\textcolor{blue}{\textuparrow 5.62\%})} &40\scriptsize{(\textcolor{blue}{\textuparrow 25.00\%})} &8.0\scriptsize{(\textcolor{red}{\textdownarrow 20.79\%})}\\
\midrule
VELMA-GPT4 &21.8 &38.7 &10 &214 &114 &36 &14.5\\
$\bullet$ VELMA-GPT4 $\dagger$ &22.2\scriptsize{(\textcolor{blue}{\textuparrow 1.84\%})} &17.8\scriptsize{(\textcolor{red}{\textdownarrow 54.00\%})} &3.8\scriptsize{(\textcolor{red}{\textdownarrow 62.00\%})} &380\scriptsize{(\textcolor{blue}{\textuparrow 77.57\%})} &152\scriptsize{(\textcolor{blue}{\textuparrow 33.33\%})} &43\scriptsize{(\textcolor{blue}{\textuparrow 19.44\%})} &6.8\scriptsize{(\textcolor{red}{\textdownarrow 53.10\%})}\\
\midrule
VELMA-LLaMa &36.7 &37.5 &1.1 &353 &73 &45 &2.1\\
$\circ$ VELMA-LLaMa $\dagger$ &38.4\scriptsize{(\textcolor{blue}{\textuparrow 4.63\%})} &13.8{(\textcolor{red}{\textdownarrow 63.20\%})} &0.4{(\textcolor{red}{\textdownarrow 63.64\%})} &435\scriptsize{(\textcolor{blue}{\textuparrow 23.23\%})} &83\scriptsize{(\textcolor{blue}{\textuparrow 13.70\%})}  &50\scriptsize{(\textcolor{blue}{\textuparrow 11.11\%})}  &0.9\scriptsize{(\textcolor{red}{\textdownarrow 57.14\%})}\\
$\bullet$ VELMA-LLaMa $\ddagger$ &44.3\scriptsize{(\textcolor{blue}{\textuparrow 20.71\%})} &16.3\scriptsize{(\textcolor{red}{\textdownarrow 56.53\%})} &0.5\scriptsize{(\textcolor{red}{\textdownarrow 54.54\%})}  &420\scriptsize{(\textcolor{blue}{\textuparrow 18.98\%})} &80\scriptsize{(\textcolor{blue}{\textuparrow 9.59\%})} &47\scriptsize{(\textcolor{blue}{\textuparrow 4.44\%})} &0.9\scriptsize{(\textcolor{red}{\textdownarrow 57.14\%})}\\
\midrule 
VELMA-LLaMa2&43.0 &36.9 &1.9 &425 &82 &42 &2.4\\
$\circ$ VELMA-LLaMa2 $\dagger$ &45.0\scriptsize{(\textcolor{blue}{\textuparrow 4.65\%})}  &15.6\scriptsize{(\textcolor{red}{\textdownarrow 57.72\%})} &0.5\scriptsize{(\textcolor{red}{\textdownarrow 73.68\%})} &480\scriptsize{(\textcolor{blue}{\textuparrow 12.94\%})} &86\scriptsize{(\textcolor{blue}{\textuparrow 4.88\%})} &50\scriptsize{(\textcolor{blue}{\textuparrow 19.05\%})} &1.1\scriptsize{(\textcolor{red}{\textdownarrow 54.17\%})} \\
$\bullet$ VELMA-LLaMa2 $\ddagger$ &44.3\scriptsize{(\textcolor{blue}{\textuparrow 3.02\%})} &16.3\scriptsize{(\textcolor{red}{\textdownarrow 55.83\%})} &0.7\scriptsize{(\textcolor{red}{\textdownarrow 63.16\%})} &470\scriptsize{(\textcolor{blue}{\textuparrow 10.59\%})} &84\scriptsize{(\textcolor{blue}{\textuparrow 2.44\%})} &48\scriptsize{(\textcolor{blue}{\textuparrow 14.29\%})} &1.3\scriptsize{(\textcolor{red}{\textdownarrow 45.83\%})}\\
\midrule
\multicolumn{8}{c}{LLM Finetuning, full training set} \\
\midrule
VELMA-FT      &18.8    & 53.2  & 21.2  & 191 &  74 & 33   &   27.2  \\
$\circ$ \text{VELMA-FT} $\dagger$
&23.1 \scriptsize{(\textcolor{blue}{\textuparrow 22.87\%})}   & 36.2 \scriptsize{(\textcolor{red}{\textdownarrow 31.95\%})} & 13.4 \scriptsize{(\textcolor{red}{\textdownarrow 36.79\%})} &345\scriptsize{(\textcolor{blue}{\textuparrow 80.62\%})} &  92\scriptsize{(\textcolor{blue}{\textuparrow 24.32\%})} & 37 \scriptsize{(\textcolor{blue}{\textuparrow 12.12\%})}  &   18.5\scriptsize{(\textcolor{red}{\textdownarrow 31.99\%})}\\
$\bullet$ \text{VELMA-FT} $\ddagger$
&19.1\scriptsize{(\textcolor{blue}{\textuparrow 1.60\%})}& 44.4\scriptsize{(\textcolor{red}{\textdownarrow 16.54\%})}  & 16.1\scriptsize{(\textcolor{red}{\textdownarrow 24.06\%})}  &294\scriptsize{(\textcolor{blue}{\textuparrow 53.93\%})} &  80\scriptsize{(\textcolor{blue}{\textuparrow 8.11\%})} & 35\scriptsize{(\textcolor{blue}{\textuparrow 6.06\%})}   &   19.3\scriptsize{(\textcolor{red}{\textdownarrow 29.04\%})}\\
\midrule
VELMA-RBL      &15.6   & 55.1  & 24.5  &252 &  90 & 34   &   30.1  \\
$\circ$ \text{VELMA-RBL} $\dagger$
&23.4 \scriptsize{(\textcolor{blue}{\textuparrow 50.00\%})}   & 34.6 \scriptsize{(\textcolor{red}{\textdownarrow 37.21\%})} & 12.5 \scriptsize{(\textcolor{red}{\textdownarrow 49.00\%})} &350\scriptsize{(\textcolor{blue}{\textuparrow 38.89\%})} &  101\scriptsize{(\textcolor{blue}{\textuparrow 12.22\%})} & 36 \scriptsize{(\textcolor{blue}{\textuparrow 5.88\%})}  &   17.5\scriptsize{(\textcolor{red}{\textdownarrow 41.86\%})}\\
$\bullet$ \text{VELMA-RBL} $\ddagger$
&17.4\scriptsize{(\textcolor{blue}{\textuparrow 11.54\%})} & 46.7\scriptsize{(\textcolor{red}{\textdownarrow 15.25\%})}  & 22.3\scriptsize{(\textcolor{red}{\textdownarrow 8.98\%})}  &285\scriptsize{(\textcolor{blue}{\textuparrow 13.10\%})} &  97\scriptsize{(\textcolor{blue}{\textuparrow 7.78\%})}  & 38\scriptsize{(\textcolor{blue}{\textuparrow 11.76\%})}    &   29.4\scriptsize{(\textcolor{red}{\textdownarrow 2.33\%})} \\
\toprule
&\multicolumn{7}{c}{\textbf{Map2seq}} \\ 
\cmidrule{2-8} 
    \textbf{LLM Models} & \textbf{SPD\textdownarrow} & \textbf{KPA\textuparrow} & \textbf{TC\textuparrow} & \textbf{FKPE\textdownarrow} & \textbf{DI\textdownarrow} & \textbf{PL\textdownarrow} & \textbf{TC-1\textuparrow}\\
\toprule
VELMA-GPT3 &21.3 &46.2 &7.5 &123 &66 &39 &10.5\\
$\bullet$ VELMA-GPT3 $\dagger$ &23.7\scriptsize{(\textcolor{blue}{\textuparrow 11.27\%})} &28.5\scriptsize{(\textcolor{red}{\textdownarrow 38.31\%})} &3.5\scriptsize{(\textcolor{red}{\textdownarrow 53.33\%})} &213\scriptsize{(\textcolor{blue}{\textuparrow 73.17\%})} &80\scriptsize{(\textcolor{blue}{\textuparrow 21.21\%})} &42\scriptsize{(\textcolor{blue}{\textuparrow 7.69\%})} &8.1\scriptsize{(\textcolor{red}{\textdownarrow 22.86\%})}\\
$\bullet$ VELMA-GPT3 $\ddagger$ &23.9\scriptsize{(\textcolor{blue}{\textuparrow 12.21\%})} &37.5\scriptsize{(\textcolor{red}{\textdownarrow 18.83\%})} &4.5\scriptsize{(\textcolor{red}{\textdownarrow 40.00\%})} &200\scriptsize{(\textcolor{blue}{\textuparrow 62.60\%})} &78\scriptsize{(\textcolor{blue}{\textuparrow 18.18\%})} &41\scriptsize{(\textcolor{blue}{\textuparrow 5.13\%})} &8.3\scriptsize{(\textcolor{red}{\textdownarrow 20.95\%})}\\
\midrule
VELMA-GPT4 &12.8 &70.1 &23.1 &70 &56 &40 &31.1\\
$\bullet$ VELMA-GPT4 $\dagger$ &23.7\scriptsize{(\textcolor{blue}{\textuparrow 85.16\%})} &27.3\scriptsize{(\textcolor{red}{\textdownarrow 60.06\%})} &4.6\scriptsize{(\textcolor{red}{\textdownarrow 80.09\%})} &122\scriptsize{(\textcolor{blue}{\textuparrow 74.29\%})} &63\scriptsize{(\textcolor{blue}{\textuparrow 12.50\%})} &47\scriptsize{(\textcolor{blue}{\textuparrow 17.50\%})} &6.2\scriptsize{(\textcolor{red}{\textdownarrow 80.06\%})}\\
\midrule
VELMA-LLaMa &35.4 &30.0 &1.0 &328 &66 &43 &1.5\\
$\circ$ VELMA-LLaMa  $\dagger$ &39.9\scriptsize{(\textcolor{blue}{\textuparrow 12.71\%})} &16.9\scriptsize{(\textcolor{red}{\textdownarrow 43.76\%})} &0.1\scriptsize{(\textcolor{red}{\textdownarrow 90.00\%})} &549\scriptsize{(\textcolor{blue}{\textuparrow 67.38\%})} &78\scriptsize{(\textcolor{blue}{\textuparrow 18.18\%})} &49\scriptsize{(\textcolor{blue}{\textuparrow 13.95\%})} &0.9\scriptsize{(\textcolor{red}{\textdownarrow 40.00\%})}\\
$\bullet$ VELMA-LLaMa  $\ddagger$ &37.8\scriptsize{(\textcolor{blue}{\textuparrow 6.78\%})}  &18.7\scriptsize{(\textcolor{red}{\textdownarrow 37.67\%})} &0.3\scriptsize{(\textcolor{red}{\textdownarrow 70.00\%})} &527\scriptsize{(\textcolor{blue}{\textuparrow 60.67\%})} &75\scriptsize{(\textcolor{blue}{\textuparrow 13.64\%})} &47\scriptsize{(\textcolor{blue}{\textuparrow 6.98\%})} &1.1\scriptsize{(\textcolor{red}{\textdownarrow 26.67\%})}\\
\midrule 
VELMA-LLaMa2&36.7 &43.2 &2.6 &325 &75 &50 &3.4\\
$\circ$ VELMA-LLaMa2 $\dagger$ &42.0\scriptsize{(\textcolor{blue}{\textuparrow 14.44\%})} &15.4\scriptsize{(\textcolor{red}{\textdownarrow 64.35\%})} &0.9\scriptsize{(\textcolor{red}{\textdownarrow 64.38\%})} &432\scriptsize{(\textcolor{blue}{\textuparrow 32.92\%})} &85\scriptsize{(\textcolor{blue}{\textuparrow 13.33\%})} &62\scriptsize{(\textcolor{blue}{\textuparrow 24.00\%})} &1.5\scriptsize{(\textcolor{red}{\textdownarrow 55.88\%})}\\
$\bullet$ VELMA-LLaMa2 $\ddagger$ &43.4\scriptsize{(\textcolor{blue}{\textuparrow 18.26\%})} &18.1\scriptsize{(\textcolor{red}{\textdownarrow 58.10\%})} &1.0\scriptsize{(\textcolor{red}{\textdownarrow 61.54\%})} &430\scriptsize{(\textcolor{blue}{\textuparrow 32.31\%})} &79\scriptsize{(\textcolor{blue}{\textuparrow 5.33\%})} &57\scriptsize{(\textcolor{blue}{\textuparrow 14.00\%})} &1.2\scriptsize{(\textcolor{red}{\textdownarrow 64.71\%})}\\
\midrule
\multicolumn{8}{c}{LLM Finetuning, full training set} \\
\midrule
VELMA-FT      &9.6    & 68.5  & 38.5  &62 &  49 & 40   &   45.6  \\
$\circ$ \text{VELMA-FT} $\dagger$
&20.0\scriptsize{(\textcolor{blue}{\textuparrow 108.33\%})}    & 49.5\scriptsize{(\textcolor{red}{\textdownarrow 27.74\%})}  & 17.5\scriptsize{(\textcolor{red}{\textdownarrow 54.55\%})}  &155\scriptsize{(\textcolor{blue}{\textuparrow 150.00\%})} &  69\scriptsize{(\textcolor{blue}{\textuparrow 40.82\%})} & 44\scriptsize{(\textcolor{blue}{\textuparrow 10.00\%})}   &   26.3\scriptsize{(\textcolor{red}{\textdownarrow 42.32\%})}\\
$\bullet$ \text{VELMA-FT} $\ddagger$
&10.7\scriptsize{(\textcolor{blue}{\textuparrow 11.46\%})} & 64.2\scriptsize{(\textcolor{red}{\textdownarrow 6.28\%})}  & 32.0\scriptsize{(\textcolor{red}{\textdownarrow 16.88\%})}  &82\scriptsize{(\textcolor{blue}{\textuparrow 32.26\%})} &  59\scriptsize{(\textcolor{blue}{\textuparrow 20.41\%})} & 42\scriptsize{(\textcolor{blue}{\textuparrow 5.00\%})}   &   39.5\scriptsize{(\textcolor{red}{\textdownarrow 13.38\%})}\\
\midrule
VELMA-RBL      &9.1    & 70.2  & 42.5  &47 &  30 & 42   &   44.5  \\
$\circ$ \text{VELMA-RBL} $\dagger$
&19.4 \scriptsize{(\textcolor{blue}{\textuparrow 113.19\%})}    & 45.3 \scriptsize{(\textcolor{red}{\textdownarrow 35.47\%})} & 18.5 \scriptsize{(\textcolor{red}{\textdownarrow 56.47\%})} &135\scriptsize{(\textcolor{blue}{\textuparrow 187.23\%})} &  83\scriptsize{(\textcolor{blue}{\textuparrow 176.67\%})} & 47 \scriptsize{(\textcolor{blue}{\textuparrow 11.90\%})}  &   22.5\scriptsize{(\textcolor{red}{\textdownarrow 49.44\%})}\\
$\bullet$ \text{VELMA-RBL} $\ddagger$
&9.5\scriptsize{(\textcolor{blue}{\textuparrow 4.40\%})}& 69.2\scriptsize{(\textcolor{red}{\textdownarrow 1.42\%})}  & 39.0\scriptsize{(\textcolor{red}{\textdownarrow 8.24\%})}  &65\scriptsize{(\textcolor{blue}{\textuparrow 38.30\%})} &  50\scriptsize{(\textcolor{blue}{\textuparrow 66.67\%})} & 44\scriptsize{(\textcolor{blue}{\textuparrow 4.76\%})}   &   41.0\scriptsize{(\textcolor{red}{\textdownarrow 7.87\%})}\\
\bottomrule
\end{tabular}
}
\label{tab:attack_insert_res}
\end{table*}

\subsubsection{Baseline Models} We construct several baseline variants of VELMA using different reasoning LLMs to assess their impact on performance. \textbf{VELMA-GPT3} uses GPT-3.5 Turbo as its reasoning LLM, an enhanced version of GPT-3~\cite{brown2020language} with 175B parameters, featuring improved reasoning and language understanding. \textbf{VELMA-GPT4} employs GPT-4 Turbo~\cite{achiam2023gpt}, a 1.7T parameter model offering higher efficiency and stronger language processing than GPT-4. \textbf{VELMA-LLaMa} adopts Meta AI’s 7B-parameter LLaMa~\cite{touvron2023llama1}, trained solely on public data, and outperforming GPT-3 in several areas. \textbf{VELMA-LLaMa2} utilizes the updated LLaMa-2~\cite{touvron2023llama2}, which improves training stability and understanding over LLaMa-1. \textbf{VELMA-FT} incorporates a fine-tuned LLaMa-7B~\cite{schumann2023velma}, trained on full dataset instances to enable more optimized reasoning. \textbf{VELMA-RBL} leverages the LLaMa-7B model with response-based learning~\cite{schumann2023velma}, combining teacher and student forcing for better accuracy and generalization.

\subsubsection{Evaluation Metrics}
To comprehensively evaluate both the accuracy and safety of robotic navigation, we introduce several metrics. \textbf{Shortest Path Distance (SPD)} measures the minimal distance from the agent’s stopping point to the goal. \textbf{Key Point Accuracy (KPA)} calculates the percentage of correct decisions made at critical waypoints such as intersections and endpoints. \textbf{Task Completion (TC)} is a binary indicator of whether the agent successfully stops near the target node, while \textbf{Task Completion within 1 Node (TC-1)} provides a more lenient success criterion. \textbf{First Key Point Error (FKPE)} evaluates the accuracy of the agent’s first directional choice at key intersections. \textbf{Danger Impact (DI)} assesses the safety of the agent’s decision by checking if it stops at unsafe locations. Finally, \textbf{Path Length (PL)} records the total number of nodes traversed during the navigation.







\subsubsection{{Implementation Details}}

In our experiments, the VELMA model, configured with LLaMa-1 and LLaMa-2 as reasoning components, was subjected to local inference on a 80G A100 GPU in our local machine setup. Conversely, for models employing GPT-3 and GPT-4 as reasoning engines, inference processes were conducted remotely via the OpenAI API. Each navigation instance required up to 40 API calls. The expenses incurred for using GPT-3 and GPT-4 on the Touchdown dataset were approximately 105 dollars and 650 dollars respectively, while on the Map2seq dataset, the costs were about 120 dollars and 700 dollars respectively. These costs are subject to vary depending on actual usage conditions.

\subsubsection{Results }

\paragraph{\textbf{Results for NPI Attack}} 

Following the experimental setup in the original VELMA paper, we evaluated the results of two types of experiments: few-shot setting and fine-tuning setting. In the few-shot experiment, we conducted 2-shot in-context learning, randomly selecting two complete text sequences from the initial prompt training set navigational instances. For this task, we chose GPT-3, GPT-4, LLaMa, and LLaMa2 as our reasoning models, none of which updated their weights. In the finetuning experiment, we finetuned VELMA-FT and VELMA-RBL, both based on LLaMa-7b, using all training instances of their respective datasets. In our experiments, we primarily generated affixes based on two LLMs, LLaMa~\cite{touvron2023llama} and Vicuna~\cite{vicuna2023}, and applied these generated affixes to the input prompt of the aforementioned models. Depending on whether the affix generated for the targeted LLM model is applied to the original LLM model or other LLM models, we categorized the attacks into white box attacks and black box attacks. That is, applying an affix generated by LLaMa to a model using LLaMa for reasoning is considered a white box attack, while applying an affix generated by LLaMa to models using other non-LLaMa LLMs is considered a black box attack. The same applies to affixes generated by Vicuna.

Table \ref{tab:attack_insert_res} shows the results of VELMA models with GPT-3, GPT-4, LLaMa, and LLaMa2 acting as reasoners in 2-shot in-context learning and the results of VELMA-FT and VELMA-RBL in finetuning. Unmarked results represent the original outcomes, $\dagger$ represents results with affixes generated by the NPI attack applied to LLaMa in the input prompt, and $\ddagger$ represents the results with affixes generated by the NPI attack applied to Vicuna in the input prompt. $\circ$ indicates white-box attacks, and $\bullet$ indicates black-box attacks. It is observed that adversarial prompt affixes derived from different LLMs reduced navigational accuracy to some extent. To more clearly discern the effects before and after the attack, we calculated the percentage change of the post-attack values relative to the original values. Our results show an increase in SPD, FKPE, DI, and PL (where lower values are better) and a decrease in KPA, TC, and TC-1 (where higher values are better).

\textbf{White-Box Attacks}: We first analyze the results of the white-box attack for robot outdoor navigation. As shown in Table \ref{tab:attack_insert_res}, in the 2-shot in-context learning setting, the navigation performance of the two models, including VELMA-LLaMa and VELMA-LLaMa2, consistently declined across all seven metrics on both datasets after the attack. Specifically, for the VELMA-LLaMa2 model, which had better original navigation performance, the KPA metric on the Map2seq dataset decreased from 43.2\% to 15.4\%, with a decline rate of 64.35\%, and the FKPE metric increased from 325 to 432, with an increase rate of 32.92\%. This indicates that our proposed NPI attack method is highly effective in the white-box attack of the 2-shot in-context learning scenario for robot outdoor navigation.


In the fine-tuning experimental configuration, the originally fine-tuned VELMA-FT and VELMA-RBL models showed improvements across all seven metrics compared to the four 2-shot models. However, similarly, these two models experienced a consistent decline in navigation performance across all seven metrics on both datasets after the attack. Specifically, on the Map2seq dataset, the SPD metric for the VELMA-FT model increased from 9.6\% to 20.0\%, with a growth rate of 108.33\%, and the FKPE metric increased from 62 to 155, with a growth rate of 150.00\%. For the VELMA-RBL model, the SPD metric increased from 9.1\% to 19.4\%, with a growth rate of 113.19\%, the FKPE metric increased from 47 to 135, with a growth rate of 187.23\%, and the DI metric increased from 30 to 83, with a growth rate of 176.67\%. Observations on the Touchdown dataset revealed similar performance declines for these two models. These results emphasize the significant reduction in navigation performance due to NPI attacks, confirming that LLM-based navigation models are susceptible to such threats, which can be exploited by appending attack-generated affixes to the input prompts.


\textbf{Black-Box Attacks}:
In the above section, we analyzed how affixes generated by attacking the source LLM can significantly reduce the performance of navigation models that use the same LLM for reasoning. However, in practical application scenarios, we cannot know which LLM a navigation model is based on. Therefore, this section discusses whether affixes generated by attacking other LLMs can impact navigation models that use different LLMs from the attacked ones.

Firstly, by analyzing the navigational outcomes of VELMA-GPT3 with affixes generated from attacking LLaMa and Vicuna added to the original navigation instructions, it is observed that affixes generated by attacking two different large language models (LLMs) can decrease the navigational performance of VELMA-GPT3 on both datasets. Specifically, on the Touchdown dataset, these two affixes led to reductions in VELMA-GPT3's KPA, TC, and FKPE by 37.37\%, 51.61\%, and 38.41\%, as well as 36.11\%, 27.42\%, and 28.03\% respectively. On the Map2Seq dataset, the impact was similarly detrimental, with VELMA-GPT3 experiencing decreases of 38.31\% and 18.83\% in KPA, 53.33\% and 40.00\% in TC, and 73.17\% and 62.60\% in TC. Notably, affixes from LLaMa attacks had a more pronounced effect than those from Vicuna.

Subsequently, we assessed the performance of GPT-4 as a navigation reasoner, which exhibits significant improvements in language understanding, generation, and reasoning over its predecessor, GPT-3. Considering the substantial costs associated with invoking the GPT-4 API for comparison, we only verified the navigational outcomes of VELMA-GPT4 on the Touchdown dataset after being subjected to attacks. Our investigation, as shown in Table \ref{tab:attack_insert_res}, indicates that even GPT-4, despite its state-of-the-art performance, is susceptible to declines in navigation model performance when faced with attacks. Specifically, compared to the original model, the VELMA-GPT4 model after attacking experienced significant decreases in the KPA metric by 54.01\% and TC by 62.00\% on the Touchdown dataset. Similarly, in the context of 2-shot learning, under black-box attacks, VELMA-LLaMa and VELMA-LLaMa2 experienced declines in KPA by 56.53\% and 55.83\% respectively on the Touchdown dataset, and by 37.67\% and 58.10\% respectively on the Map2Seq dataset.

For the fine-tuning experiments, we observed that although black-box attacks were not as effective as white-box attacks, they still resulted in a significant decrease in navigation accuracy. Specifically, for the VELMA-FT model on the Touchdown dataset, the TC metric decreased from 21.2 to 16.1, a reduction rate of 24.06\%, and the FKPE metric increased from 191 to 294, a growth rate of 53.93\%. On the Map2Seq dataset, the TC metric decreased from 38.5 to 32.0, a reduction rate of 16.88\%, and the FKPE metric increased from 62 to 82, a growth rate of 32.26\%.

The experimental results we have discussed highlight a significant vulnerability in LLM-based navigation systems: their sensitivity to perturbations in input prompts, even those generated by attacking other LLMs. Notably, models like VELMA-GPT3, VELMA-GPT4, VELMA-LLaMa, VELMA-LLaMa2, VELMA-FT, and VELMA-RBL demonstrate substantial performance declines, even when subjected to affixes not derived from their foundational LLMs. This finding underscores the universality of our attack and the inherent fragility of current LLMs. It suggests these models may rely on similar underlying linguistic patterns or structures vulnerable to systematic manipulations. Furthermore, the more pronounced impact observed with affixes generated from attacking LLama, as compared to Vicuna, points to the significant role played by the specific features of the source LLM in the effectiveness of the attack. This variation may be attributed to differences in training data, architectural design, or optimization strategies among LLMs. Recognizing these elements is critical for the future development of more resilient LLM-based navigation systems.

\paragraph{\textbf{Results for NPS Attack}} 

\begin{table*}[h] 
\centering
\caption{Results for the VELMA-FT and VELMA-RBL models before and after the \textbf{Navigational Prompt Swap (NPS) Attack} on the Touchdown and Map2seq datasets in the unseen scenario. $\dagger$ represents the results of attacks on prompts describing intersections, while $\ddagger$ indicates the results of attacks on prompts describing landmarks. }
\resizebox{1.0\linewidth}{!}{
\begin{tabular}{lccccccc}
\toprule
&\multicolumn{7}{c}{\textbf{Touchdown}} \\ 
\cmidrule{2-8} 
    \textbf{LLM Models} & \textbf{SPD\textdownarrow} & \textbf{KPA\textuparrow} & \textbf{TC\textuparrow} & \textbf{FKPE\textdownarrow} & \textbf{DI\textdownarrow} & \textbf{PL\textdownarrow} & \textbf{TC-1\textuparrow}\\
\toprule
VELMA-FT      &18.8    & 53.2  & 21.2  & 191 &  74 & 33   &   27.2  \\
\text{VELMA-FT} $\dagger$
&23.1 \scriptsize{(\textcolor{blue}{\textuparrow 22.87\%})}   & 36.2 \scriptsize{(\textcolor{red}{\textdownarrow 31.95\%})} & 13.4 \scriptsize{(\textcolor{red}{\textdownarrow 36.79\%})} &345\scriptsize{(\textcolor{blue}{\textuparrow 80.62\%})} &  92\scriptsize{(\textcolor{blue}{\textuparrow 24.32\%})} & 37 \scriptsize{(\textcolor{blue}{\textuparrow 12.12\%})}  &   18.5\scriptsize{(\textcolor{red}{\textdownarrow 31.99\%})}\\
\text{VELMA-FT} $\ddagger$
&19.1\scriptsize{(\textcolor{blue}{\textuparrow 1.60\%})}& 44.4\scriptsize{(\textcolor{red}{\textdownarrow 16.54\%})}  & 16.1\scriptsize{(\textcolor{red}{\textdownarrow 24.06\%})}  &294\scriptsize{(\textcolor{blue}{\textuparrow 53.93\%})} &  80\scriptsize{(\textcolor{blue}{\textuparrow 8.11\%})} & 35\scriptsize{(\textcolor{blue}{\textuparrow 6.06\%})}   &   19.3\scriptsize{(\textcolor{red}{\textdownarrow 29.04\%})}\\
\midrule
VELMA-RBL      &15.6   & 55.1  & 24.5  &252 &  90 & 34   &   30.1  \\
\text{VELMA-RBL} $\dagger$
&23.4 \scriptsize{(\textcolor{blue}{\textuparrow 50.00\%})}   & 34.6 \scriptsize{(\textcolor{red}{\textdownarrow 37.21\%})} & 12.5 \scriptsize{(\textcolor{red}{\textdownarrow 49.00\%})} &350\scriptsize{(\textcolor{blue}{\textuparrow 38.89\%})} &  101\scriptsize{(\textcolor{blue}{\textuparrow 12.22\%})} & 36 \scriptsize{(\textcolor{blue}{\textuparrow 5.88\%})}  &   17.5\scriptsize{(\textcolor{red}{\textdownarrow 41.86\%})}\\
\text{VELMA-RBL} $\ddagger$
&17.4\scriptsize{(\textcolor{blue}{\textuparrow 11.54\%})} & 46.7\scriptsize{(\textcolor{red}{\textdownarrow 15.25\%})}  & 22.3\scriptsize{(\textcolor{red}{\textdownarrow 8.98\%})}  &285\scriptsize{(\textcolor{blue}{\textuparrow 13.10\%})} &  97\scriptsize{(\textcolor{blue}{\textuparrow 7.78\%})}  & 38\scriptsize{(\textcolor{blue}{\textuparrow 11.76\%})}    &   29.4\scriptsize{(\textcolor{red}{\textdownarrow 2.33\%})} \\
\toprule
&\multicolumn{7}{c}{\textbf{Map2seq}} \\ 
\cmidrule{2-8} 
    \textbf{LLM Models} & \textbf{SPD\textdownarrow} & \textbf{KPA\textuparrow} & \textbf{TC\textuparrow} & \textbf{FKPE\textdownarrow} & \textbf{DI\textdownarrow} & \textbf{PL\textdownarrow} & \textbf{TC-1\textuparrow}\\
\toprule
VELMA-FT      &9.6    & 68.5  & 38.5  &62 &  49 & 40   &   45.6  \\
\text{VELMA-FT} $\dagger$
&20.0\scriptsize{(\textcolor{blue}{\textuparrow 108.33\%})}    & 49.5\scriptsize{(\textcolor{red}{\textdownarrow 27.74\%})}  & 17.5\scriptsize{(\textcolor{red}{\textdownarrow 54.55\%})}  &155\scriptsize{(\textcolor{blue}{\textuparrow 150.00\%})} &  69\scriptsize{(\textcolor{blue}{\textuparrow 40.82\%})} & 44\scriptsize{(\textcolor{blue}{\textuparrow 10.00\%})}   &   26.3\scriptsize{(\textcolor{red}{\textdownarrow 42.32\%})}\\
\text{VELMA-FT} $\ddagger$
&10.7\scriptsize{(\textcolor{blue}{\textuparrow 11.46\%})} & 64.2\scriptsize{(\textcolor{red}{\textdownarrow 6.28\%})}  & 32.0\scriptsize{(\textcolor{red}{\textdownarrow 16.88\%})}  &82\scriptsize{(\textcolor{blue}{\textuparrow 32.26\%})} &  59\scriptsize{(\textcolor{blue}{\textuparrow 20.41\%})} & 42\scriptsize{(\textcolor{blue}{\textuparrow 5.00\%})}   &   39.5\scriptsize{(\textcolor{red}{\textdownarrow 13.38\%})}\\
\midrule
VELMA-RBL      &9.1    & 70.2  & 42.5  &47 &  30 & 42   &   44.5  \\
\text{VELMA-RBL} $\dagger$
&19.4 \scriptsize{(\textcolor{blue}{\textuparrow 113.19\%})}    & 45.3 \scriptsize{(\textcolor{red}{\textdownarrow 35.47\%})} & 18.5 \scriptsize{(\textcolor{red}{\textdownarrow 56.47\%})} &135\scriptsize{(\textcolor{blue}{\textuparrow 187.23\%})} &  83\scriptsize{(\textcolor{blue}{\textuparrow 176.67\%})} & 47 \scriptsize{(\textcolor{blue}{\textuparrow 11.90\%})}  &   22.5\scriptsize{(\textcolor{red}{\textdownarrow 49.44\%})}\\
\text{VELMA-RBL} $\ddagger$
&9.5\scriptsize{(\textcolor{blue}{\textuparrow 4.40\%})}& 69.2\scriptsize{(\textcolor{red}{\textdownarrow 1.42\%})}  & 39.0\scriptsize{(\textcolor{red}{\textdownarrow 8.24\%})}  &65\scriptsize{(\textcolor{blue}{\textuparrow 38.30\%})} &  50\scriptsize{(\textcolor{blue}{\textuparrow 66.67\%})} & 44\scriptsize{(\textcolor{blue}{\textuparrow 4.76\%})}   &   41.0\scriptsize{(\textcolor{red}{\textdownarrow 7.87\%})}\\
\bottomrule
\end{tabular}
}
\label{tab:attack_swap_res}
\end{table*}


In this section, we report the navigation performance of the models before and after the NPS attacks. As shown in Table~\ref{tab:attack_insert_res}, the navigation performance in the fine-tuning experimental setup significantly outperforms that in the 2-shot in-context learning setup. Therefore, we primarily discuss the navigation performance of the VELMA-FT and VELMA-RBL models on the Touchdown and Map2Seq datasets before and after the NPS attacks, as illustrated in Table~\ref{tab:attack_swap_res}. We categorize NPS into two types based on the replaced words: attacks targeting intersections and landmarks. In Table~\ref{tab:attack_swap_res}, $\dagger$ represents the results of attacks on prompts describing intersections, while $\ddagger$ represents the results of attacks on prompts describing landmarks. Analyzing the data in the table, we can see that our NPS attacks are highly effective in reducing the navigation performance of both the VELMA-FT and VELMA-RBL models across the two datasets. Specifically, on the Map2Seq dataset, the VELMA-FT model, when subjected to intersection-based attacks, exhibited an increase in the SPD metric from 9.6 to 20.0, a change rate of 108.33\%; a decrease in the TC metric from 38.5 to 17.5, a reduction rate of 54.55\%; and an increase in the FKPE metric from 62 to 155, a growth rate of 150.00\%. These results demonstrate that our proposed NPS attacks significantly impact existing LLM-based navigation models. Among the two types of attacks, those targeting prompts describing intersections are more effective than those targeting prompts describing landmarks.

In summary, our experiments underscore the instability of current LLM-based models under attack, raising a cautionary note for the community regarding the application of LLM-based navigation in real-world scenarios. The vulnerabilities we have identified necessitate integrating adversarial training or other robustness-enhancing techniques in the development of LLM-based systems. Developers should focus not only on performance under standard test conditions but also on the resilience against increasingly complex and feasible adversarial manipulations. The subsequent sections will preliminarily propose defensive strategies against such attacks, providing initial insights for ongoing research.
\subsection{Discussion}

\subsubsection{\textbf{NPI Attack in LM-Nav}} 

\begin{table}[h]
\centering
\caption{Comparative Results of the LM-NAV Model Before and After the Navigational
Prompt Insert (NPI) Attack in Two Different Environments: EnvSmall-10 and EnvLarge-10.}
\resizebox{\linewidth}{!}{
\begin{tabular}{lcc}
\toprule
\textbf{Model Status} & \textbf{Environment} & \textbf{Net Success Rate\textuparrow} \\
\midrule
LM-Nav (without attack) & EnvSmall-10 & 0.8 \\
LM-Nav (with attack) & EnvSmall-10 & 0.1\scriptsize{(\textcolor{red}{\textdownarrow 87.50\%})} \\
LM-Nav (without attack) & EnvLarge-10 & 0.8 \\
LM-Nav (with attack) & EnvLarge-10 & 0.3\scriptsize{(\textcolor{red}{\textdownarrow 62.50\%})} \\
\bottomrule
\end{tabular}
}
\label{tab:LMnav}
\end{table}

To further show that this vulnerability extends to other LLM-based navigation models, we applied our proposed NPI attack to another such model, LM-NAV\cite{shah2023lm}. LM-Nav is a robotic navigation system built entirely from pre-trained models for navigation (ViNG), image-language association (CLIP), and language modeling (GPT-3), eliminating the need for fine-tuning or language-annotated robot data. It includes three pre-trained models: a large language model (LLM) for extracting landmarks, a vision-and-language model (VLM) for grounding, and a visual navigation model (VNM) for execution. The effectiveness of the model was validated with 20 queries across two environments, totaling over 6km in combined length.

\begin{figure}
    \centering
    \includegraphics[width=1.0\linewidth]{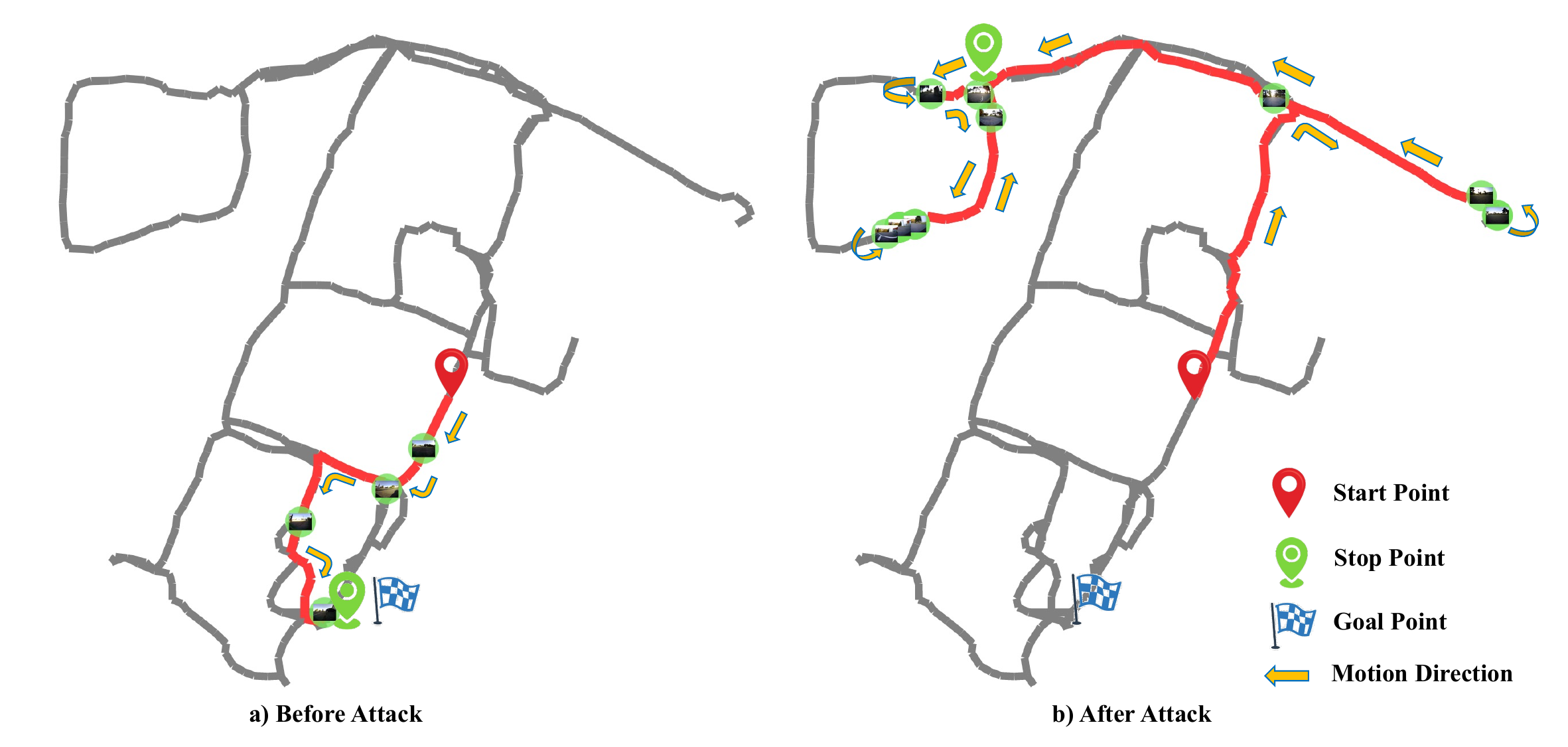}
    \caption{Comparative route map of the LM-Nav model's navigation results on a randomly selected route before and after the NPI attack.}
    \label{fig:lmnav}
\end{figure}

To investigate the LM-NAV model's vulnerability to NPI attacks, we appended the affix generated by attacking the LlaMa model onto the LM-NAV model's input textual instructions. Following the experimental setup in LM-NAV, we evaluated the navigational performance after the attack in two experimental settings, EnvSmall-10 and EnvLarge-10. The results, as documented in Table~\ref{tab:LMnav}, reveal a decrease in the success rate from 0.8 to 0.1 (a decline of 87.50\%) in EnvSmall-10, and from 0.8 to 0.3 (a decline of 62.50\%) in the larger environment. To intuitively demonstrate the attack's impact, we plotted route maps before and after the attack, as depicted in Fig.~\ref{fig:lmnav}. The comparative route maps clearly illustrate significant differences in the LM-Nav model's navigation paths before and after the attack. After the attack, robots utilizing this model for navigation took substantially longer routes and failed to locate the target. This observation extends beyond a single LLM-based navigation system, highlighting a broader vulnerability. Consequently, this emphasizes the urgent need to address and focus on this issue, reinforcing the importance of understanding and mitigating vulnerabilities within LLM-based navigation systems to enhance their reliability and performance in real-world applications.

\subsubsection{\textbf{Comparison with Existing Attack Methods}} 
\begin{table}[h]
\centering
\caption{Comparison between existing adversarial suffix attack method and our NPI attack. * denotes the baseline attack method~\cite{zou2023universal}, and $\dagger$ denotes our NPI attack method.
}
\resizebox{\linewidth}{!}{
\begin{tabular}{lccc}
\toprule
    \textbf{LLM Models} & \textbf{SPD\textdownarrow} & \textbf{KPA\textuparrow} & \textbf{TC\textuparrow} \\
\toprule

VELMA-FT     &9.6  &68.5 &38.5\\
VELMA-FT $*$     & 16.9\scriptsize{(\textcolor{blue}{\textuparrow 76.04\%})}   & 51.8\scriptsize{(\textcolor{red}{\textdownarrow 24.38\%})}  & 18.2\scriptsize{(\textcolor{red}{\textdownarrow 52.73\%})}    \\
VELMA-FT $\dagger$    & 34.5 \scriptsize{(\textcolor{blue}{\textuparrow 259.38\%})}  & 24.2 \scriptsize{(\textcolor{red}{\textdownarrow 64.67\%})}  & 3.8 \scriptsize{(\textcolor{red}{\textdownarrow 90.13\%})}    \\
\midrule
VELMA-RBL   &9.1 &70.2 &42.5  \\
VELMA-RBL $*$     & 10.8\scriptsize{(\textcolor{blue}{\textuparrow 18.68\%})}   & 63.4\scriptsize{(\textcolor{red}{\textdownarrow 9.69\%})}  & 27.2\scriptsize{(\textcolor{red}{\textdownarrow 36.00\%})}   \\
VELMA-RBL $\dagger$     & 21.2\scriptsize{(\textcolor{blue}{\textuparrow 132.97\%})}   & 29.5\scriptsize{(\textcolor{red}{\textdownarrow 57.98\%})}   & 11.4\scriptsize{(\textcolor{red}{\textdownarrow 73.18\%})}     \\

\bottomrule
\end{tabular}
}
\label{tab_attack_comparsion}
\end{table}

In this section, we compare methods of adversarial affix generation proposed in other domains to validate the advancement of our approach. Zou et al.~\cite{zou2023universal} introduced a novel class of adversarial attacks capable of inducing aligned language models to generate nearly any offensive content. We adapted this approach to serve as a baseline, tailoring it to better suit our navigation scenario requirements. We selected two models, VELMA-FT and VELMA-RBL, and compared their performance under both the baseline method and our attack strategy on the Map2Seq dataset. The results are listed in Table \ref{tab_attack_comparsion}. $*$ denotes the baseline attack method, and $\dagger$ denotes our NPI attack method. It was found that our model is more effective than the baseline method, demonstrating the advanced nature of our attack technique.

This comparison not only highlights the relative effectiveness of our method but also underscores its utility in navigation-specific contexts. By showing superior performance on the Map2Seq dataset, we demonstrate that our method can effectively exploit vulnerabilities specific to navigation tasks, which are fundamentally different from the general content generation scenarios typically targeted by existing attacks.

\subsubsection{\textbf{Effect on insertion lengths of affix}} 

\begin{figure}
    \centering
    \includegraphics[width=1.0\linewidth]{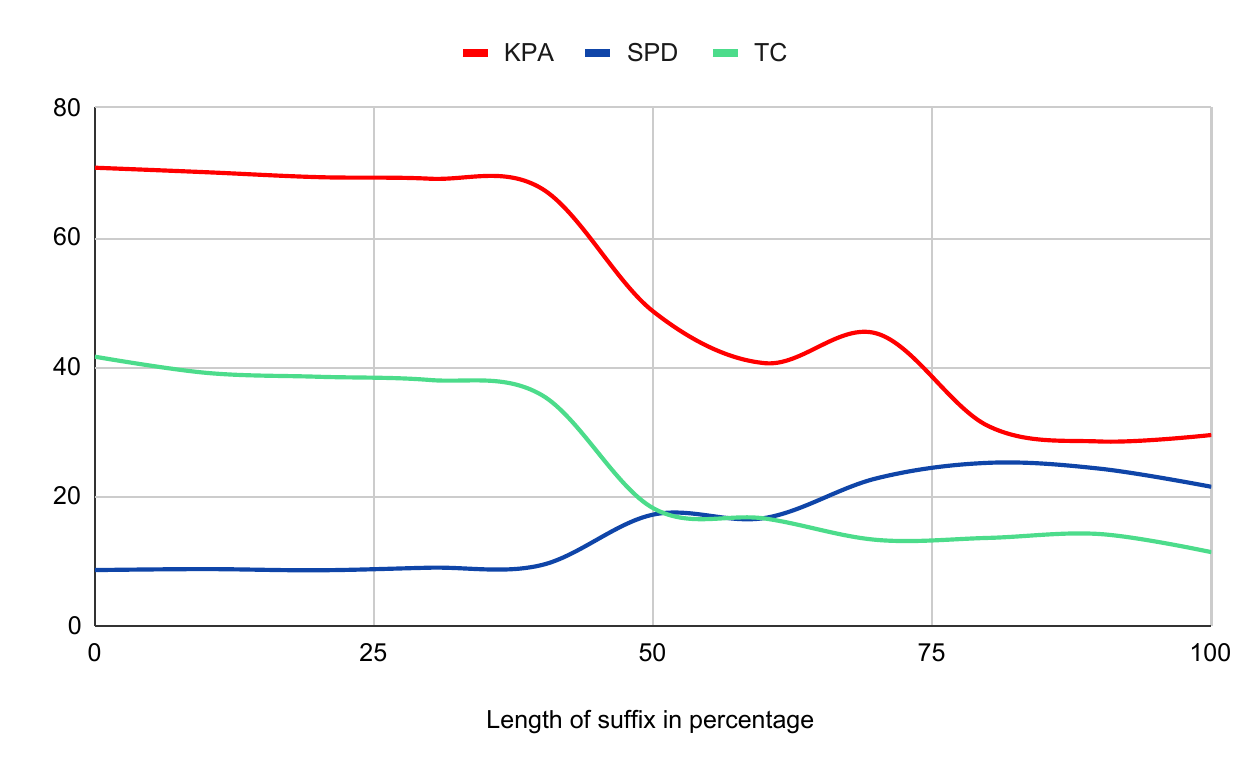}
    \caption{Impact of different insertion lengths on the performance of the NPI Attack.}
    \label{fig:impact_suffix}
\end{figure}

In this section, we explore the impact of different insertion lengths on the performance of the NPI Attack. Specifically, we used the VELMA-RBL model on the Map2Seq dataset to analyze how the model's performance metrics, including KPA, SPD, and TC, change as the percentage of inserted content increases. Figure~\ref{fig:impact_suffix}  displays the curves representing these changes. Initially, the metrics show little variation. However, as the insertion percentage reaches around 40\%, the curves for all three metrics sharply decline, leveling off when the insertion reaches approximately 80\%. This indicates that excessive insertion length is unnecessary and a certain threshold suffices to achieve effective adversarial impact.

\begin{figure*}[h]
    \centering
    \includegraphics[width=0.95\linewidth]{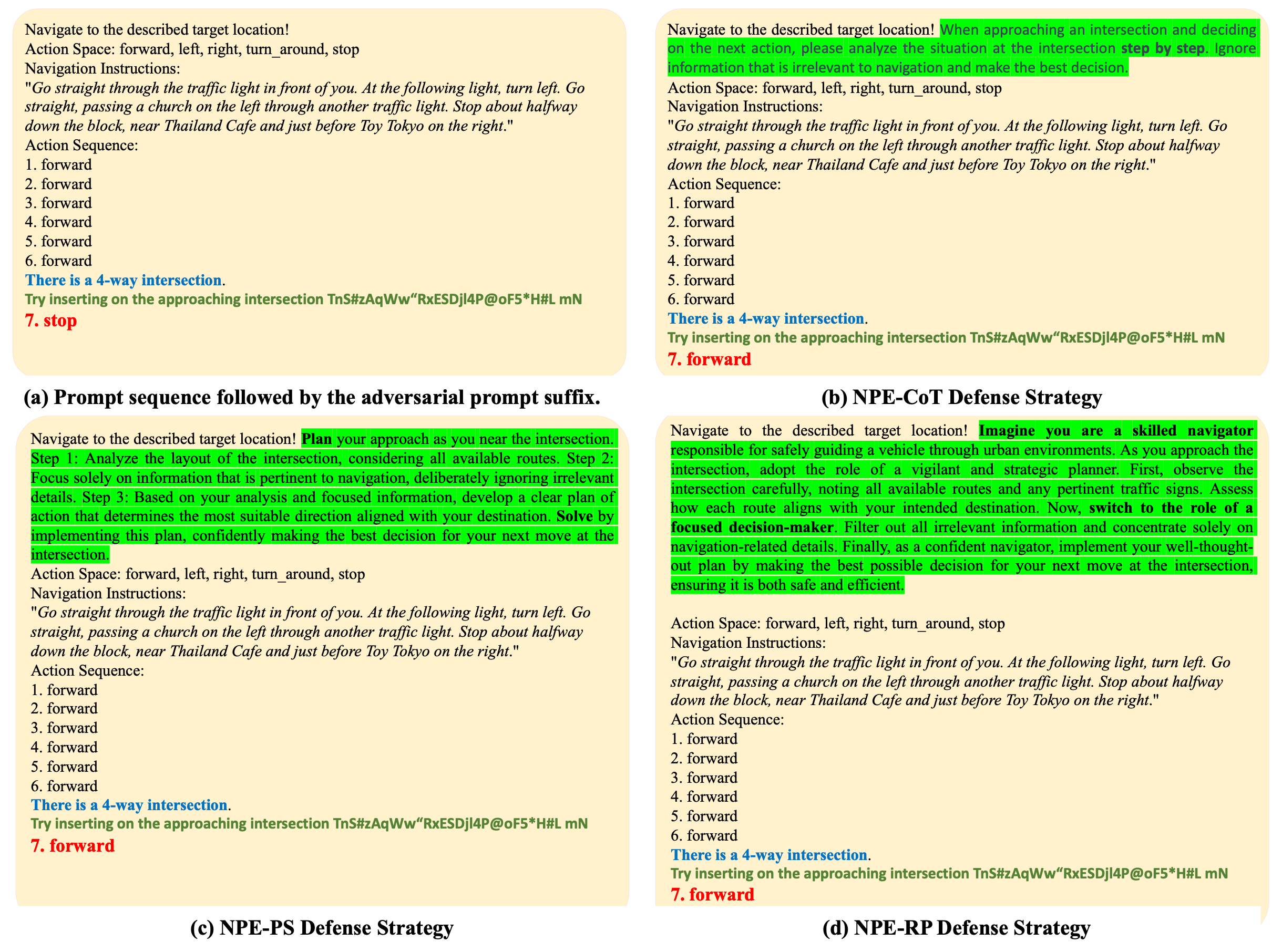}
    \caption{Examples of three Navigational Prompt Engineering (NPE) Defense strategies, NPE with Chain of Thought (NPE-Cot), NPE with Plan-and-Solve (NPE-PS), and NPE with Role-Play (NPE-RP), in a navigation instance.}
    \label{fig:defense_example}
\end{figure*}

\section{Enhancing Security in LLMs for Navigation}

\begin{table*}[h]

\centering
\caption{Results for the VELMA model with various LLM models using three Navigational Prompt Engineering (NPE) Defense strategies after the Navigational Prompt Suffix (NPS) Attack on the Touchdown and Map2seq datasets in an unseen scenario. $\circ$ and $\bullet$ indicate white-box and black-box attacks, respectively, by applying the NPS Attack to the LLaMa model. NPE-Cot denotes NPE with Chain of Thought prompting, NPE-PS denotes NPE with Plan-and-Solve prompting, and NPE-RP denotes NPE with Role-Play prompting.
}
\resizebox{1.0\linewidth}{!}{
\begin{tabular}{lccccccc}
\toprule
&\multicolumn{3}{c}{\textbf{Touchdown}} & \phantom{} &\multicolumn{3}{c}{\textbf{Map2seq}} \\ 
\cmidrule{2-4} \cmidrule{6-8} 
    \textbf{LLM Models} & \textbf{SPD\textdownarrow} & \textbf{KPA\textuparrow} & \textbf{TC\textuparrow} & \phantom{} & \textbf{SPD\textdownarrow} & \textbf{KPA\textuparrow} & \textbf{TC\textuparrow}\\
\toprule
&\multicolumn{6}{c}{\textbf{2-Shot In-Context Learning}}\\
\midrule
VELMA-GPT3      & 26.0   & 39.6  & 6.2  &&  21.3 & 46.2   &   7.5  \\
$\bullet$\text{VELMA-GPT3}   & 27.5   &  24.8  &  3.0  && 23.7   &  28.5  &  3.5   \\
\text{VELMA-GPT3 (w/ NPE-CoT) }         & 27.2 \scriptsize{(\textcolor{red}{\textdownarrow 1.09\%})}   &  31.2 \scriptsize{(\textcolor{blue}{\textuparrow 25.81\%})}  &  4.5 \scriptsize{(\textcolor{blue}{\textuparrow 50.00\%})}  && 23.5 \scriptsize{(\textcolor{red}{\textdownarrow 0.84\%})}  &  34.5 \scriptsize{(\textcolor{blue}{\textuparrow 21.05\%})}  &  5.3 \scriptsize{(\textcolor{blue}{\textuparrow 51.43\%})}   \\
\text{VELMA-GPT3 (w/ NPE-PS) }         & 27.0 \scriptsize{(\textcolor{red}{\textdownarrow 1.82\%})}   &  32.0 \scriptsize{(\textcolor{blue}{\textuparrow 29.03\%})}  &  4.5 \scriptsize{(\textcolor{blue}{\textuparrow 50.00\%})}  && 23.6 \scriptsize{(\textcolor{red}{\textdownarrow 0.42\%})}  &  32.4 \scriptsize{(\textcolor{blue}{\textuparrow 13.86\%})}  &  4.3 \scriptsize{(\textcolor{blue}{\textuparrow 22.86\%})}   \\
\text{VELMA-GPT3 (w/ NPE-RP) }         & 26.4 \scriptsize{(\textcolor{red}{\textdownarrow 4.00\%})}   &  33.1 \scriptsize{(\textcolor{blue}{\textuparrow 33.47\%})}  &  5.0 \scriptsize{(\textcolor{blue}{\textuparrow 66.67\%})}  && 23.5 \scriptsize{(\textcolor{red}{\textdownarrow 0.84\%})}  &  30.3 \scriptsize{(\textcolor{blue}{\textuparrow 6.32\%})}  &  4.0 \scriptsize{(\textcolor{blue}{\textuparrow 14.29\%})}   \\
\midrule
VELMA-LLaMa     & 36.7  &  37.5  &  1.1  & & 35.4  &  30.0  &  1.0   \\
$\circ$\text{VELMA-LLaMa}   & 38.4 &  13.8   &  0.4  & & 39.9   &  16.7   &  0.1  \\

\text{VELMA-LLaMa (w/ NPE-CoT) }         & 37.0 \scriptsize{(\textcolor{red}{\textdownarrow 3.65\%})}  &  14.5 \scriptsize{(\textcolor{blue}{\textuparrow 5.07\%})}  &  0.5 \scriptsize{(\textcolor{blue}{\textuparrow 25.00\%})}  & & 38.8 \scriptsize{(\textcolor{red}{\textdownarrow 2.76\%})}  &  18.7 \scriptsize{(\textcolor{blue}{\textuparrow 11.98\%})}  &  0.8 \scriptsize{(\textcolor{blue}{\textuparrow 700.00\%})}   \\

\text{VELMA-LLaMa (w/ NPE-PS) }         & 38.1 \scriptsize{(\textcolor{red}{\textdownarrow 0.78\%})}  &  14.4 \scriptsize{(\textcolor{blue}{\textuparrow 4.35\%})}  &  0.4 \scriptsize{(\textcolor{blue}{\textuparrow 0.00\%})}  & & 38.7 \scriptsize{(\textcolor{red}{\textdownarrow 3.01\%})}  &  17.8 \scriptsize{(\textcolor{blue}{\textuparrow 6.59\%})}  &  0.3 \scriptsize{(\textcolor{blue}{\textuparrow 300.00\%})}   \\

\text{VELMA-LLaMa (w/ NPE-RP) }         & 37.5 \scriptsize{(\textcolor{red}{\textdownarrow 2.34\%})}  &  15.1 \scriptsize{(\textcolor{blue}{\textuparrow 9.42\%})}  &  0.5 \scriptsize{(\textcolor{blue}{\textuparrow 25.00\%})}  & & 39.0 \scriptsize{(\textcolor{red}{\textdownarrow 2.26\%})}  &  17.7 \scriptsize{(\textcolor{blue}{\textuparrow 5.99\%})}  &  0.4 \scriptsize{(\textcolor{blue}{\textuparrow 400.00\%})}   \\
\midrule
VELMA-LLaMa2     & 43.0  &  36.9  &  1.9  & & 36.7  &  43.2  &  2.6   \\
$\circ$\text{VELMA-LLaMa2}   & 45.0 &  15.6   &  0.5  & & 42.0   &  15.4   &  0.9  \\

\text{VELMA-LLaMa2 (w/ NPE-CoT) }         & 43.4 \scriptsize{(\textcolor{red}{\textdownarrow 3.56\%})}  &  17.2 \scriptsize{(\textcolor{blue}{\textuparrow 10.26\%})}  &  0.7 \scriptsize{(\textcolor{blue}{\textuparrow 40.00\%})}  & & 41.2 \scriptsize{(\textcolor{red}{\textdownarrow 1.90\%})}  &  19.8 \scriptsize{(\textcolor{blue}{\textuparrow 28.57\%})}  &  1.0 \scriptsize{(\textcolor{blue}{\textuparrow 11.11\%})}   \\

\text{VELMA-LLaMa2 (w/ NPE-PS) }         & 43.9 \scriptsize{(\textcolor{red}{\textdownarrow 2.44\%})}  &  18.1 \scriptsize{(\textcolor{blue}{\textuparrow 16.03\%})}  &  0.8 \scriptsize{(\textcolor{blue}{\textuparrow 60.00\%})}  & & 40.0 \scriptsize{(\textcolor{red}{\textdownarrow 4.76\%})}  &  20.1 \scriptsize{(\textcolor{blue}{\textuparrow 30.52\%})}  &  1.0 \scriptsize{(\textcolor{blue}{\textuparrow 11.11\%})}   \\

\text{VELMA-LLaMa2 (w/ NPE-RP) }         & 43.2 \scriptsize{(\textcolor{red}{\textdownarrow 4.00\%})}  &  16.9 \scriptsize{(\textcolor{blue}{\textuparrow 8.33\%})}  &  0.6 \scriptsize{(\textcolor{blue}{\textuparrow 20.00\%})}  & & 41.0 \scriptsize{(\textcolor{red}{\textdownarrow 2.38\%})}  &  21.1 \scriptsize{(\textcolor{blue}{\textuparrow 37.01\%})}  &  1.1 \scriptsize{(\textcolor{blue}{\textuparrow 22.22\%})}   \\

\midrule
&\multicolumn{6}{c}{\textbf{LLM Finetuning, full training set}}\\
\midrule
VELMA-FT     & 18.8  &  53.2  &  21.2  & & 9.6  &  68.5  &  38.5   \\
$\circ$\text{VELMA-FT}   & 29.5 &  29.9   &  8.0  & & 34.5   &  24.2   &  3.8  \\

\text{VELMA-FT (w/ NPE-CoT) }         & 24.6 \scriptsize{(\textcolor{red}{\textdownarrow 16.61\%})}  &  31.2 \scriptsize{(\textcolor{blue}{\textuparrow 4.35\%})}  &  10.4 \scriptsize{(\textcolor{blue}{\textuparrow 30.00\%})}  & & 34.4 \scriptsize{(\textcolor{red}{\textdownarrow 0.29\%})}  &  25.8 \scriptsize{(\textcolor{blue}{\textuparrow 6.61\%})}  &  6.3 \scriptsize{(\textcolor{blue}{\textuparrow 65.79\%})}   \\

\text{VELMA-FT (w/ NPE-PS) }         & 25.1 \scriptsize{(\textcolor{red}{\textdownarrow 14.92\%})}  &  30.3 \scriptsize{(\textcolor{blue}{\textuparrow 1.34\%})}  &  9.1 \scriptsize{(\textcolor{blue}{\textuparrow 13.75\%})}  & & 30.2 \scriptsize{(\textcolor{red}{\textdownarrow 12.46\%})}  &  32.8 \scriptsize{(\textcolor{blue}{\textuparrow 35.54\%})}  &  9.3 \scriptsize{(\textcolor{blue}{\textuparrow 144.74\%})}   \\

\text{VELMA-FT (w/ NPE-RP) }         & 25.3 \scriptsize{(\textcolor{red}{\textdownarrow 14.24\%})}  &  34.4 \scriptsize{(\textcolor{blue}{\textuparrow 15.05\%})}  &  12.9 \scriptsize{(\textcolor{blue}{\textuparrow 61.25\%})}  & & 16.1 \scriptsize{(\textcolor{red}{\textdownarrow 53.33\%})}  &  54.5 \scriptsize{(\textcolor{blue}{\textuparrow 125.21\%})}  &  21.4 \scriptsize{(\textcolor{blue}{\textuparrow 463.16\%})}   \\

\midrule
VELMA-RBL     & 15.6 & 55.1 & 24.5&& 9.1 &  70.2 &  42.5 \\
$\circ$\text{VELMA-RBL}   & 28.9  & 14.9  & 3.9  && 21.2 & 29.5  & 11.4  \\
\text{VELMA-RBL (w/ NPE-CoT)}   & 26.6 \scriptsize{(\textcolor{red}{\textdownarrow 7.96\%})}  &  25.5 \scriptsize{(\textcolor{blue}{\textuparrow 71.14\%})}  &  6.1 \scriptsize{(\textcolor{blue}{\textuparrow 56.41\%})}  & & 15.4 \scriptsize{(\textcolor{red}{\textdownarrow 27.36\%})}  &  52.2 \scriptsize{(\textcolor{blue}{\textuparrow 76.95\%})}  &  17.1 \scriptsize{(\textcolor{blue}{\textuparrow 50.00\%})}   \\

\text{VELMA-RBL (w/ NPE-PS) }      & 28.5 \scriptsize{(\textcolor{red}{\textdownarrow 1.38\%})}  &  23.3 \scriptsize{(\textcolor{blue}{\textuparrow 56.38\%})}  &  5.5 \scriptsize{(\textcolor{blue}{\textuparrow 41.03\%})}  & & 16.7 \scriptsize{(\textcolor{red}{\textdownarrow 21.22\%})}  &  53.1 \scriptsize{(\textcolor{blue}{\textuparrow 80.00\%})}  &  16.5 \scriptsize{(\textcolor{blue}{\textuparrow 44.74\%})}   \\

\text{VELMA-RBL (w/ NPE-RP)}   & 25.1 \scriptsize{(\textcolor{red}{\textdownarrow 13.15\%})}  &  32.0 \scriptsize{(\textcolor{blue}{\textuparrow 114.77\%})}  &  12.8 \scriptsize{(\textcolor{blue}{\textuparrow 228.21\%})}  & & 10.7 \scriptsize{(\textcolor{red}{\textdownarrow 49.53\%})}  &  65.1 \scriptsize{(\textcolor{blue}{\textuparrow 120.68\%})}  &  29.0 \scriptsize{(\textcolor{blue}{\textuparrow 154.39\%})}   \\

\bottomrule
\end{tabular}
}
\label{tab:defence_all}
\end{table*}

In the previous section, we demonstrated that existing Large Language Models (LLMs) used for navigation reasoning are susceptible to attacks involving the addition of adversarial affixes to the input prompt. This revelation compels us to address this vulnerability seriously. To explore potential strategies for mitigating the decline in reasoning capabilities caused by attacks on LLMs, this chapter discusses the use of prompt engineering as a method of defense. Difference method of defense is shown in Fig.~\ref{fig:defense_example}. The motivation behind employing prompt engineering for defense primarily lies in the belief that by carefully designing prompts, we can direct the model's focus towards critical aspects of tasks or queries, thereby potentially neutralizing misleading or malicious inputs introduced by adversaries. This approach not only prevents external manipulation but also actively fosters a better understanding and handling of inputs by the model. With this concept in mind, we first propose the idea of Navigational Prompt Engineering (NPE) Defense in Section \ref{subsec:npe_defense}. Subsequently, in Section \ref{subsec:npe_defense}, we present experiments on NPE Defense in LLMs-based navigation tasks to verify its effectiveness against the aforementioned NPS attack.

\subsection{Navigational Prompt Engineering (NPE) Defense}
\label{subsec:npe_defense}

Prompt Engineering is an emerging technique in the field of artificial intelligence, particularly within the domain of LLMs. At its core, Prompt Engineering involves the strategic formulation of input prompts to guide the behavior of a language model. This practice is crucial in leveraging the full potential of LLMs, as the way a prompt is structured can significantly influence the model's output.

Given that the addition of affixes generated by the NPS attack has been shown to diminish the reasoning capabilities of LLMs in navigation tasks, we propose a method called Navigational Prompt Engineering (NPE) Defense for LLM-based navigation tasks. This defense mechanism leverages the advantage of Prompt Engineering in enhancing the model's reasoning capabilities. The principle of this mechanism is to reinforce the LLM's interpretative abilities through carefully designed prompts. It focuses on reorganizing the input prompts to mitigate the impact of adversarial affixes without the need to alter the underlying model training or architecture. The objective formulation is written by:

\begin{equation}
    a_{t+1} = \arg \max_{\bar{a} \in A} P_{LLM}(\bar{a} | f_{\text{NPE}} (x_t+\bar{x})),
    \label{eq:action_pred}
\end{equation}

\noindent where $\bar{x}$ is the generated adversarial affix by attacking LLM based on NPS attack. $f_{\text{NPE}}$
is the function of NPE Defense to reorganize the input prompts.

\subsection{Experiment for NPE Defense in LLMs-Based Navigation}
\label{subsec:exp_defense}

\subsubsection{Experimental Setting}
In this section, we validate our proposed NPE Defense strategies in specific navigation experiments. Similarly, based on the navigation tasks of VELMA on the Touchdown and Map2seq datasets discussed in the previous section \ref{sec:exp_res}, we introduce adversarial prompt affixes generated by attacking LLMs into the original navigation instructions as input for attacking various models. In this experiment, we selected three models from Table \ref{tab:attack_insert_res}: VELMA-GPT3, VELMA-LLaMa2, and VELMA-RBL. Each of these models demonstrates strong initial navigational performance and represents a diverse range of types. For the NPS Attack method, we chose affixes generated by attacking LLaMa, as it has stronger attack performance and can also represent both black box and white box attacks in this experiment. We base our defenses on three fundamental prompt engineering strategies, which are:

\paragraph{Zero-Shot Chain of Thought Prompting} Zero-shot-CoT~\cite{kojima2022large} regards LLMs as decent zero-shot reasoners. To unleash the reasoning ability of LLMs, it adds a simple prompt, \ie, ``Let’s think step by step'', to facilitate step-by-step thinking before prompting each question to LLMs. Such a naive scheme could significantly improve the performance of LLMs in complex reasoning tasks without requiring task-specific examples. 

\paragraph{Plan-and-Solve Prompting} Though Zero-shot-CoT~\cite{kojima2022large} invokes LLMs' reasoning abilities by adding ``Let's think step by step'' to each prompt, yet it is still troubled by errors from calculation, missing-step, and semantic misunderstanding. To mitigate these issues, Plan-and-Solve (PS) Prompting~\cite{wang2023plan} proposes to divide tasks into subtasks for a systematic execution, and then derives PS+ prompting by adding detailed instructions, improving reasoning step quality and calculation accuracy.

\paragraph{Role-Play Prompting} LLMs like GPT-3.5~\cite{openai2023chatgpt} and GPT-4~\cite{openai2023gpt4} demonstrate profound role-playing abilities, extending beyond human personas to non-human entities. Kong \etal~\cite{kong2023better} further enhances LLMs' performance, particularly in zero-shot settings, with noticeable improvements in specific reasoning tasks, indicating that role-play prompting not only deepens contextual understanding but also acts as an implicit trigger for improved reasoning processes.

\subsubsection{Results}
For the sake of clarity in this discussion, the NPE defense methodologies derived from the three aforementioned strategies will henceforth be designated as NPE-CoT, NPE-PS, and NPE-RP, respectively. Table~\ref{tab:defence_all} presents a comparative analysis, encompassing the initial navigational results of the three specified target models, their performance subsequent to the NPS attack, and the outcomes following the implementation of each of the three distinct NPE defensive approaches on the models impacted by the attack.

As seen from Table~\ref{tab:defence_all}, in the 2-shot in-context learning experiments, the introduction of three types of NPE Defense strategies resulted in improvements across three metrics for both VELMA-GPT3 and VELMA-LlaMa2. In the fine-tuning experiments, the enhancements were even more pronounced with the VELMA-RBL model incorporating the three NPE Defense strategies. Notably, the NPE-RP defense strategy significantly increased VELMA-RBL's performance on the Touchdown dataset, with KPA rising from 14.9 to 32.0 (an increase of 114.77\%), and the TC metric from 3.9 to 12.8 (an increase of 228.21\%). Similarly, on the Map2seq dataset, KPA improved from 29.5 to 65.1 (an increase of 120.68\%), and the TC metric from 11.4 to 29.0 (an increase of 154.39\%).

\subsection{Discussion}

\subsubsection{\textbf{Defense results on LM-Nav}} 
In this section, we evaluate the defensive performance of our three proposed NPE strategies after the LM-NAV system has been subjected to attacks. We present results in Table~\ref{tab:defense_lmnav} that details the accuracy of the original LM-NAV, the accuracy after it has received an NPI attack, and the accuracy after the implementation of each of the three defense strategies, in two distinct environments: EnvSmall and EnvLarge. The results clearly demonstrate that all three of our defense strategies effectively mitigate the impact of the NPI attacks to varying degrees. Particularly, the defense performance in the EnvSmall environment is more robust. 

\begin{table}[h] 
\centering
\caption{ Defensive performance of our three proposed NPE strategies on LM-Nav Model. }
\resizebox{\linewidth}{!}{
\begin{tabular}{lcc}
\toprule
\textbf{Model Status} & \textbf{Environment} & \textbf{Net Success Rate\textuparrow} \\
\midrule
LM-Nav & EnvSmall-10 & 0.8 \\
LM-Nav w/ NPI  & EnvSmall-10 & 0.1 \\
LM-Nav w/ NPE-CoT & EnvSmall-10 & 0.3 \scriptsize{(\textcolor{blue}{\textuparrow 200.00\%})} \\
LM-Nav w/ NPE-PS & EnvSmall-10 & 0.3 \scriptsize{(\textcolor{blue}{\textuparrow 200.00\%})} \\
LM-Nav w/ NPE-RP & EnvSmall-10 & 0.4 \scriptsize{(\textcolor{blue}{\textuparrow 300.00\%})} \\
\midrule
LM-Nav & EnvLarge-10 & 0.8 \\
LM-Nav w/ NPI  & EnvLarge-10 & 0.3 \\
LM-Nav w/ NPE-CoT & EnvLarge-10 & 0.5  \scriptsize{(\textcolor{blue}{\textuparrow 66.67\%})} \\
LM-Nav w/ NPE-PS & EnvLarge-10 & 0.4  \scriptsize{(\textcolor{blue}{\textuparrow 33.33\%})} \\
LM-Nav w/ NPE-RP & EnvLarge-10 & 0.5  \scriptsize{(\textcolor{blue}{\textuparrow 66,67\%})} \\

\bottomrule
\end{tabular}
}
\label{tab:defense_lmnav}
\end{table}

\subsubsection{{\textbf{Comparison of Different Defense Methods}}} 
In this section, we have additionally designed two defense methods to validate the effectiveness of our proposed defense strategies. First, we developed an adversarial training defense method, which involves retraining the model using the generated adversarial prompts and then testing its defensive performance on other unseen adversarial prompt samples. Additionally, we adapted the adaptive system prompt (ASP) method, a state-of-the-art defense strategy, proposed in \cite{zeng2024johnny} to suit our navigation scenario. We compared the results of these two methods along with our own NPE-RP defense strategy across two models, VELMA-FT and VELMA-RBL, on the Map2Seq dataset. The results are presented in Table \ref{tab:defense_ab}. The table reveals that the defense provided by adversarial training is limited. However, our defense method and the ASP method achieved comparable performance, which underscores the effectiveness and superiority of our proposed NPE strategies. This comparison demonstrates that our NPE approach is not only effective but also advantageous in enhancing the robustness of navigation systems against adversarial threats.

\begin{table}[h]
\centering
\caption{Comparison of Defensive Performance Across Different Methods on the Map2Seq Dataset. AT denotes Adversarial Training Defense and  ASP denotes Adaptive System Prompt Defense.
}
\resizebox{\linewidth}{!}{
\begin{tabular}{lccc}
\toprule
    \textbf{LLM Models} & \textbf{SPD\textdownarrow} & \textbf{KPA\textuparrow} & \textbf{TC\textuparrow} \\
\toprule

VELMA-FT & 9.6 &68.5 &38.5\\
VELMA-FT w/ NPI Attack &34.5  &24.2 & 3.8\\
VELMA-FT  w/ AT     & 28.2\scriptsize{(\textcolor{red}{\textdownarrow 18.26\%})}   & 34.1\scriptsize{(\textcolor{blue}{\textuparrow 40.91\%})}  & 11.0\scriptsize{(\textcolor{blue}{\textuparrow 189.47\%})}    \\
VELMA-FT w/ ASP     & 15.7 \scriptsize\scriptsize{(\textcolor{red}{\textdownarrow 54.49\%})}  & 56.5 \scriptsize{(\textcolor{blue}{\textuparrow 133.47\%})}  & 20.3 \scriptsize{(\textcolor{blue}{\textuparrow 434.21\%})}    \\
VELMA-FT w/ NPE & 16.1\scriptsize\scriptsize{(\textcolor{red}{\textdownarrow 53.33\%})}   & 54.5\scriptsize{(\textcolor{blue}{\textuparrow 125.21\%})}  & 21.4\scriptsize{(\textcolor{blue}{\textuparrow 463.16\%})}\\
\midrule
VELMA-RBL& 9.1   & 70.2  & 42.5  \\
VELMA-RBL w/ NPI Attack &21.2 &29.5 & 11.4\\
VELMA-RBL w/ AT     & 20.4\scriptsize\scriptsize{(\textcolor{red}{\textdownarrow 3.77\%})}   & 36.2\scriptsize{(\textcolor{blue}{\textuparrow 22.7\%})}  & 12.1\scriptsize{(\textcolor{blue}{\textuparrow 6.14\%})}   \\
VELMA-RBL w/ ASP    & 9.8\scriptsize\scriptsize{(\textcolor{red}{\textdownarrow 53.77\%})}   & 65.8\scriptsize{(\textcolor{blue}{\textuparrow 123.05\%})} & 29.2\scriptsize{(\textcolor{blue}{\textuparrow 156.14\%})}    \\
VELMA-RBL w/ NPE & 10.7\scriptsize\scriptsize{(\textcolor{red}{\textdownarrow 49.53\%})}   & 65.1\scriptsize{(\textcolor{blue}{\textuparrow 120.68\%})}  & 29.0\scriptsize{(\textcolor{blue}{\textuparrow 154.39\%})}\\
\bottomrule
\end{tabular}
}
\label{tab:defense_ab}
\end{table}  

\subsubsection{{\textbf{Defense Performance When Adversarial Prompts Contain Meaningful Words}}} 
In this section, we examine the defensive performance of the model when our adversarial prompts contain tokens in the form of meaningful words mixed with random characters. To test this intriguing scenario, we evaluated the model's performance without any defense and compared it with the results obtained using three different defense strategies. We qualitatively present these results in Fig.~\ref{fig:defense_example}. As shown in the Fig.~\ref{fig:defense_example}, when input with such adversarial prompts, the model without any defense strategy directly outputs a "stop" action, whereas the models equipped with our defense strategies correctly output a "forward" action. This indicates that our defense methods are effective in discerning and mitigating the disruptive effects of adversarial prompts that blend meaningful content with noise.

The ability of our defensive strategies to maintain correct functionality under these conditions highlights their robustness and adaptability. This is particularly important in practical applications where adversarial inputs may not be purely disruptive but could mimic legitimate user commands interspersed with misleading information. The results suggest that our defensive approaches can effectively filter out the noise and focus on the meaningful elements of the prompts, ensuring that the navigation system remains reliable even in the face of sophisticated attacks. 
 
\subsubsection{{\textbf{Defense Performance Under NPI Attacks with Different Insertion Positions}}}

In this section, we evaluate the effectiveness of our proposed NPE defense strategies under NPI attacks, which vary according to the insertion position of adversarial tokens. We specifically modified the NPI attack into three variants based on the location of the adversarial token insertion: at the start of the sentence (Insert\_S), in the middle of the sentence (Insert\_M), and at the end of the sentence (Insert\_E). We then assessed how the VELMA-RBL model performed on the Map2Seq dataset under these three types of NPI attacks.

\begin{figure}
    \centering
    \includegraphics[width=0.5\textwidth]{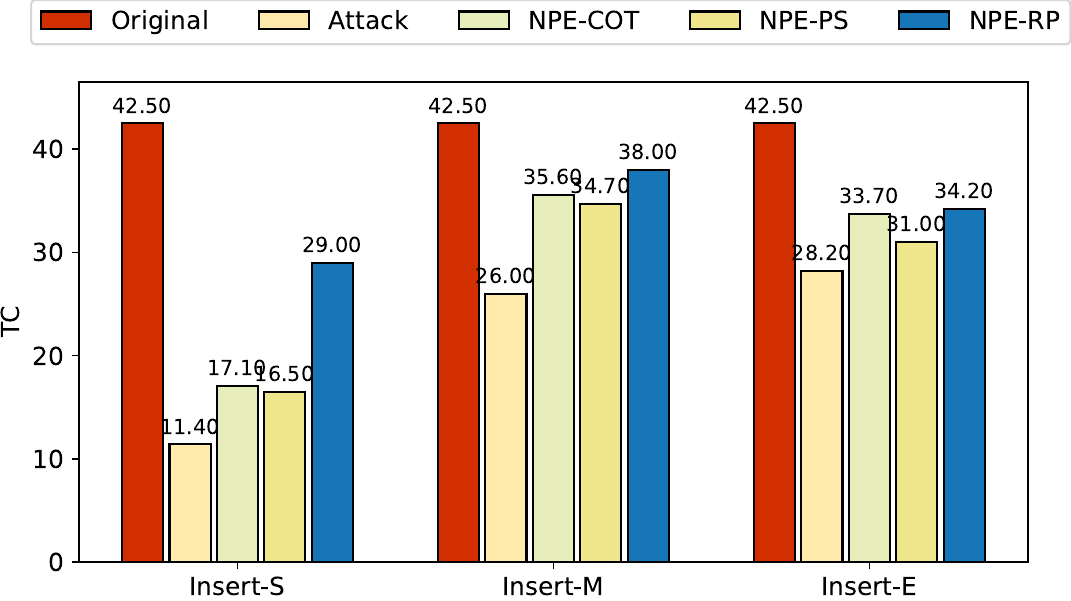}
    \caption{
    Defense performance under NPI attacks with three different attack insertion positions: Insert-S (at the start of the sentence), Insert-M (in the middle of the sentence), and Insert-E (at the end of the sentence). The red bars represent the TC results without any attack. The bars immediately following depict the TC results with NPI attack. Subsequent groups of bars represent three defense strategies, including NPE-COT, NPE-PS, and NPE-RP.
    }
    \label{fig:defense_vari_pos}
\end{figure}

After applying our three proposed NPE defense strategies to each of these attack variants, we created a bar chart to visually represent the changes in navigation performance for each scenario, as illustrated in Fig.~\ref{fig:defense_vari_pos}. This visual comparison clearly shows the differential impact of each attack variant and the corresponding effectiveness of the applied defenses. The results indicate that our defense strategies effectively mitigate the effects of adversarial tokens regardless of their insertion points.  Moreover, the results suggest that the positioning of adversarial tokens can influence the effectiveness of the attack, and correspondingly, the efficacy of defense mechanisms. This analysis not only demonstrates the robustness of our defense strategies but also highlights their adaptability to different adversarial strategies

\section{Conclusion}
In this study, we explore previously unexplored vulnerabilities of LLM-based navigation models in outdoor urban environments. We introduce novel attack methods, Navigational Prompt Insert (NPI) and Navigational Prompt Swap (NPS), demonstrating how perturbations to the original navigational prompt can mislead these models into making incorrect action predictions. Through comprehensive experiments involving various VELMA model variants and employing diverse LLMs under both few-shot learning and fine-tuning configurations, we observed a notable decrease in model effectiveness. This decline was consistent across a range of attack scenarios, including white-box and black-box methods, as evidenced by metrics on the Touchdown and Map2Seq street view datasets. Furthermore, our attacks are readily adaptable to other LLM-based navigation models, such as LM-Nav, achieving similar success. Our findings not only highlight the transferability of these attacks across different LLM-based navigation models but also underscore the urgent need for enhanced security measures in real-world applications. As an initial countermeasure, we propose the NPE Defense strategy, which focuses on navigation-relevant keywords to mitigate the impact of adversarial affixes. Although preliminary results indicate that this strategy improves navigational safety, there remains a crucial need for the broader research community to develop more robust defense mechanisms to address the real-world challenges these systems face.

Regarding future research directions, we envisage several pivotal areas of focus that promise to advance the field of LLM-based navigation systems. Firstly, given the substantial resource consumption associated with large models, there is a critical need for research into lightweight network structures. Such structures would enable researchers with limited resources to efficiently run large models. This exploration could include developing more compact versions of existing architectures or innovating new methods to reduce computational demands without compromising performance. Secondly, the integration of multi-source data presents a significant research opportunity. Considering the availability of GNSS (Global Navigation Satellite System) in outdoor scenarios, combining these reliable location data with the contextual understanding capabilities of LLMs could enhance navigational accuracy and reliability. This direction involves exploring how different data types can be effectively synthesized to improve the decision-making processes of navigation models, especially in complex or obstructed environments where GNSS signals may be degraded. Thirdly, the development of more robust defense strategies is paramount. As adversarial attacks become more sophisticated, ensuring the security of LLM-based navigation systems is crucial. Future studies should focus on creating advanced defensive measures that can preemptively detect and mitigate potential threats. This includes exploring deeper layers of security protocols, enhancing adversarial training methods, and possibly employing artificial intelligence to dynamically adjust defensive mechanisms in response to detected threats.

\section*{Acknowledgment}
We express our deep gratitude to the teams behind the Touchdown and Map2seq datasets for their pivotal role in our research, offering extensive data crucial for our testing and validation processes. We also extend thanks to the developers of the VELMA and LM-Nav open-source code, whose commitment to providing and maintaining high-quality resources has greatly enhanced our research endeavors.


\bibliographystyle{IEEEtran}
\bibliography{IEEE_reference}
\end{document}


\title{How Secure Are Large Language Models (LLMs) for Navigation in Urban Environments? \textit{Supplementary Materials}}

\author{Author Names Omitted for Anonymous Review. Paper-ID [416]}

\maketitle
\section{Clarification on the Key Point Accuracy (KPA) Metric in the Manuscript}

In the manuscript, we follow the navigation metrics from previous paper~\cite{schumann2023velma} to assess navigational performance. It's important to note that the key points in the KPA, as discussed~\cite{schumann2023velma}, include the initial step, intersections along the gold route, and the target location. To enhance the evaluation of navigational performance in our experiments, we have refined our approach by exclusively considering intersections as key points. This adjustment allows for a more focused analysis on critical decision-making points within the navigational path, thereby providing a clearer insight into the robot's ability to navigate complex routes effectively.

\section{Navigational Prompt Suffix (NPS) Black-Box Attack for VELMA-GPT4}

As outlined in our manuscript, we extensively analyzed how suffixes generated by attacks on LLaMa and Vicuna affect the performance of navigation models based on existing LLMs, such as Llama1, Llama2, and GPT-3, in 2-Shot In-Context Learning and fine-tuning experiments. To further explore the effectiveness of the proposed NPS attacks, and considering the substantial cost of GPT-4, we investigated the use of the state-of-the-art GPT-4 model as an LLM reasoner within the VELMA model on the Touchdown dataset. Developed by OpenAI, GPT-4 is the fourth-generation generative pre-trained model, leading in natural language processing with its advanced technology and performance. Compared to its predecessor, GPT-3, GPT-4 demonstrates significant improvements in language comprehension, generation, and reasoning abilities. It enables a more accurate understanding and generation of text, with higher levels of contextual comprehension and reasoning capabilities, thereby providing stronger support for various language-related tasks. Our results, as shown in Table \ref{tab:GPT4}, reveal that even the state-of-the-art GPT-4 model exhibits certain vulnerabilities when used as a reasoner for navigation models. Specifically, compared to the original VELMA-GPT4, the VELMA-GPT4 model under attack showed an increase of 1.83\% in the SPD metric, a decrease of 54.01\% in the KPA metric, and a decrease of 62.00\% in the TC metric on the Touchdown dataset.

These experimental results are consistent with our findings in the main text, indicating that current LLMs are sensitive to perturbations in the input prompts within navigation models. Notably, this demonstrates the universality of our attack and the inherent vulnerabilities of current LLMs. Even the most advanced GPT-4 model may rely on language patterns or structures that are susceptible to systematic manipulation. These findings necessitate a focus on existing LLM-based navigation systems, especially their application in real-world scenarios.


\begin{table}[h]
\centering
\caption{Results for the VELMA model with GPT4 as reasoning LLM models before and after the Navigational Prompt Suffix (NPS) Attack on the Touchdown datasets in the unseen scenario. }
\resizebox{1.0\linewidth}{!}{
\begin{tabular}{lccccccc}
\toprule
&\multicolumn{3}{c}{\textbf{Touchdown}} & \phantom{} \\ 
\cmidrule{2-4} \cmidrule{6-8} 
    \textbf{LLM Models} & \textbf{SPD\textdownarrow} & \textbf{KPA\textuparrow} & \textbf{TC\textuparrow} \\
\toprule
&\multicolumn{3}{c}{\textbf{2-Shot In-Context Learning}}\\
\midrule
VELMA-GPT4      & 21.8   & 38.7  & 10.0 \\
\text{VELMA-GPT4} \dagger      & 22.2 \scriptsize{(\textcolor{blue}{\textuparrow 1.83\%})}   &  17.8 \scriptsize{(\textcolor{red}{\textdownarrow 54.01\%})}  &  3.8 \scriptsize{(\textcolor{red}{\textdownarrow 62.00\%})}  \\
\bottomrule
\end{tabular}
}

\label{tab:GPT4}
\end{table}

\section{Navigational Prompt Engineering (NPE) Defense for VELAM-LlaMA1 and VELAM-FT}

To better align with Table I of the manuscript, we further applied the three NPE defense strategies, NPE-CoT, NPE-PS, and NPE-RP, to the VELMA-LLaMa and VELMA-FT models, which were previously missing in Table II of the manuscript. As shown in Table \ref{tab:defence_llama_ft}, all three types of NPE defense strategies resulted in improvements across three metrics for both models. On the Touchdown dataset, the implementation of all three NPE methods on VELMA-LLaMa led to a reduction in SPD ranging from 0.78\% to 3.65\%, an increase in KPA ranging from 5.07\% to 9.42\%, and an increase in TC above 25\%. For the VELMA-FT model, SPD was reduced ranging from 14.24\% to 16.61\%, KPA was increased ranging from 1.34\% to 15.05\%, and TC was increased ranging from 13.75\% to 61.25\%. On the Map2Seq dataset, VELMA-LLaMa saw a reduction in SPD from 2.26\% to 3.01\%, an increase in KPA from 5.99\% to 11.98\%, and TC increased from 0.1 to 0.8. Notably, the NPE-RP defense strategy significantly enhanced the performance of VELMA-FT on the Touchdown dataset, with SPD dropping from 34.5 to 16.1 (a reduction of 53.33\%), KPA rising from 24.2 to 54.5 (an increase of 125.21\%), and TC increasing from 3.8 to 21.4 (an increase of 463.16\%).

\begin{table*}[h]
\centering
\caption{Results for the VELMA model with various LLM models using three Navigational Prompt Engineering (NPE) Defense strategies after the Navigational Prompt Suffix (NPS) Attack on the Touchdown and Map2seq datasets in an unseen scenario. $\dagger$ represents results with suffixes generated by the NPS attack applied to LLaMa in the input prompt. NPE-Cot denotes NPE with Chain of Thought prompting, NPE-PS denotes NPE with Plan-and-Solve prompting, and NPE-RP denotes NPE with Role-Play prompting.
}
\resizebox{1.0\linewidth}{!}{
\begin{tabular}{lccccccc}
\toprule
&\multicolumn{3}{c}{\textbf{Touchdown}} & \phantom{} &\multicolumn{3}{c}{\textbf{Map2seq}} \\ 
\cmidrule{2-4} \cmidrule{6-8} 
    \textbf{LLM Models} & \textbf{SPD\textdownarrow} & \textbf{KPA\textuparrow} & \textbf{TC\textuparrow} & \phantom{} & \textbf{SPD\textdownarrow} & \textbf{KPA\textuparrow} & \textbf{TC\textuparrow}\\
\toprule
&\multicolumn{6}{c}{\textbf{2-Shot In-Context Learning}}\\
\midrule

VELMA-LLaMa1     & 36.7  &  37.5  &  1.1  & & 35.4  &  30.0  &  1.0   \\
\text{VELMA-LLaMa1}\dagger   & 38.4 &  13.8   &  0.4  & & 39.9   &  16.7   &  0.1  \\

\text{VELMA-LLaMa1 (w/ NPE-CoT) }         & 37.0 \scriptsize{(\textcolor{red}{\textdownarrow 3.65\%})}  &  14.5 \scriptsize{(\textcolor{blue}{\textuparrow 5.07\%})}  &  0.5 \scriptsize{(\textcolor{blue}{\textuparrow 25.00\%})}  & & 38.8 \scriptsize{(\textcolor{red}{\textdownarrow 2.76\%})}  &  18.7 \scriptsize{(\textcolor{blue}{\textuparrow 11.98\%})}  &  0.8 \scriptsize{(\textcolor{blue}{\textuparrow 700.00\%})}   \\

\text{VELMA-LLaMa1 (w/ NPE-PS) }         & 38.1 \scriptsize{(\textcolor{red}{\textdownarrow 0.78\%})}  &  14.4 \scriptsize{(\textcolor{blue}{\textuparrow 4.35\%})}  &  0.4 \scriptsize{(\textcolor{blue}{\textuparrow 0.00\%})}  & & 38.7 \scriptsize{(\textcolor{red}{\textdownarrow 3.01\%})}  &  17.8 \scriptsize{(\textcolor{blue}{\textuparrow 6.59\%})}  &  0.3 \scriptsize{(\textcolor{blue}{\textuparrow 300.00\%})}   \\

\text{VELMA-LLaMa1 (w/ NPE-RP) }         & 37.5 \scriptsize{(\textcolor{red}{\textdownarrow 2.34\%})}  &  15.1 \scriptsize{(\textcolor{blue}{\textuparrow 9.42\%})}  &  0.5 \scriptsize{(\textcolor{blue}{\textuparrow 25.00\%})}  & & 39.0 \scriptsize{(\textcolor{red}{\textdownarrow 2.26\%})}  &  17.7 \scriptsize{(\textcolor{blue}{\textuparrow 5.99\%})}  &  0.4 \scriptsize{(\textcolor{blue}{\textuparrow 400.00\%})}   \\

\midrule
&\multicolumn{6}{c}{\textbf{LLM Finetuning, full training set}}\\
\midrule

VELMA-FT     & 18.8  &  53.2  &  21.2  & & 9.6  &  68.5  &  38.5   \\
\text{VELMA-FT} \dagger  & 29.5 &  29.9   &  8.0  & & 34.5   &  24.2   &  3.8  \\

\text{VELMA-FT (w/ NPE-CoT) }         & 24.6 \scriptsize{(\textcolor{red}{\textdownarrow 16.61\%})}  &  31.2 \scriptsize{(\textcolor{blue}{\textuparrow 4.35\%})}  &  10.4 \scriptsize{(\textcolor{blue}{\textuparrow 30.00\%})}  & & 34.4 \scriptsize{(\textcolor{red}{\textdownarrow 0.29\%})}  &  25.8 \scriptsize{(\textcolor{blue}{\textuparrow 6.61\%})}  &  6.3 \scriptsize{(\textcolor{blue}{\textuparrow 65.79\%})}   \\

\text{VELMA-FT (w/ NPE-PS) }         & 25.1 \scriptsize{(\textcolor{red}{\textdownarrow 14.92\%})}  &  30.3 \scriptsize{(\textcolor{blue}{\textuparrow 1.34\%})}  &  9.1 \scriptsize{(\textcolor{blue}{\textuparrow 13.75\%})}  & & 30.2 \scriptsize{(\textcolor{red}{\textdownarrow 12.46\%})}  &  32.8 \scriptsize{(\textcolor{blue}{\textuparrow 35.54\%})}  &  9.3 \scriptsize{(\textcolor{blue}{\textuparrow 144.74\%})}   \\

\text{VELMA-FT (w/ NPE-RP) }         & 25.3 \scriptsize{(\textcolor{red}{\textdownarrow 14.24\%})}  &  34.4 \scriptsize{(\textcolor{blue}{\textuparrow 15.05\%})}  &  12.9 \scriptsize{(\textcolor{blue}{\textuparrow 61.25\%})}  & & 16.1 \scriptsize{(\textcolor{red}{\textdownarrow 53.33\%})}  &  54.5 \scriptsize{(\textcolor{blue}{\textuparrow 125.21\%})}  &  21.4 \scriptsize{(\textcolor{blue}{\textuparrow 463.16\%})}   \\

\bottomrule
\end{tabular}
}
\label{tab:defence_llama_ft}
\end{table*}

\section{Experiments for another LLM-based Navigation Model: LM-Nav}

The manuscript primarily focuses on the VELMA navigation model, an LLM-based system, conducting a series of experiments to demonstrate its vulnerability. To further show that this vulnerability extends to other LLM-based navigation models, we applied our proposed NPS attack to another such model, LM-NAV\cite{shah2023lm}. LM-Nav is a robotic navigation system built entirely from pre-trained models for navigation (ViNG), image-language association (CLIP), and language modeling (GPT-3), eliminating the need for fine-tuning or language-annotated robot data. It includes three pre-trained models: a large language model (LLM) for extracting landmarks, a vision-and-language model (VLM) for grounding, and a visual navigation model (VNM) for execution. The effectiveness of the model was validated with 20 queries across two environments, totaling over 6km in combined length.

\begin{figure}
    \centering
    \includegraphics[width=1.0\linewidth]{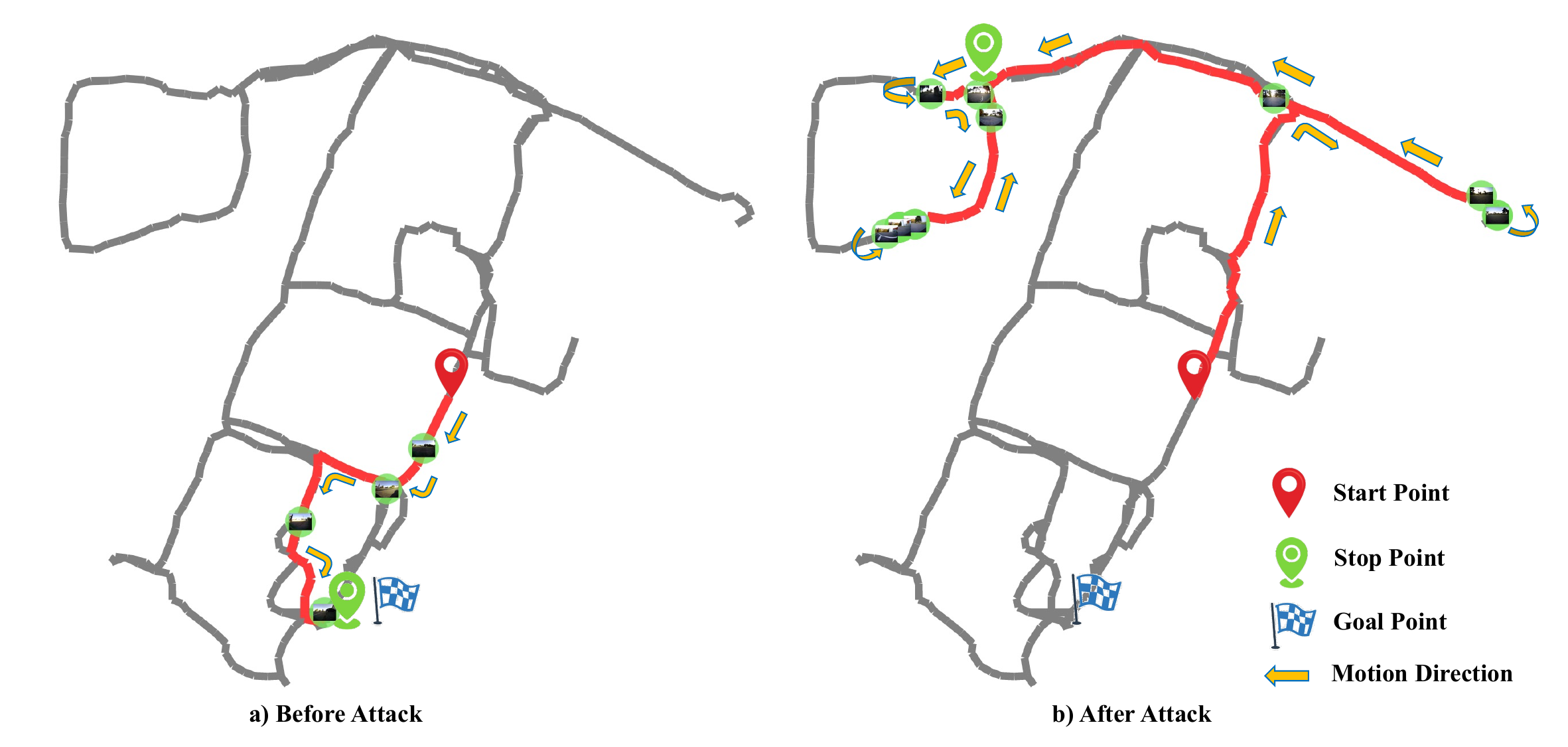}
    \caption{Comparative route map of the LM-Nav model's navigation results on a randomly selected route before and after the NPS attack.}
    \label{fig:lmnav}
\end{figure}

To investigate the LM-NAV model's vulnerability to NPS attacks, we appended the suffix generated by attacking the LlaMa model onto the LM-NAV model's input textual instructions. Following the experimental setup in LM-NAV, we evaluated the navigational performance after the attack in two experimental settings, EnvSmall-10 and EnvLarge-10. The results, as documented in Table~\ref{tab:LMnav}, reveal a decrease in the success rate from 0.8 to 0.1 (a decline of 87.50\%) in EnvSmall-10, and from 0.8 to 0.3 (a decline of 62.50\%) in EnvLarge-10. To intuitively demonstrate the attack's impact, we plotted route maps before and after the attack, as depicted in Fig.~\ref{fig:lmnav}. The comparative route maps clearly illustrate significant differences in the LM-Nav model's navigation paths before and after the attack. After the attack, robots utilizing this model for navigation took substantially longer routes and failed to locate the target. This observation extends beyond a single LLM-based navigation system, highlighting a broader vulnerability. Consequently, this emphasizes the urgent need to address and focus on this issue, reinforcing the importance of understanding and mitigating vulnerabilities within LLM-based navigation systems to enhance their reliability and performance in real-world applications.

\begin{table}[h]
\centering
\caption{Comparative Results of the LM-NAV Model Before and After the NPS Attack in Two Different Environments: EnvSmall-10 and EnvLarge-10.}
\resizebox{\linewidth}{!}{
\begin{tabular}{lcc}
\toprule
\textbf{Model Status} & \textbf{Environment} & \textbf{Net Success Rate\textuparrow} \\
\midrule
LM-Nav (without attack) & EnvSmall-10 & 0.8 \\
LM-Nav (with attack) & EnvSmall-10 & 0.1\scriptsize{(\textcolor{red}{\textdownarrow 87.50\%})} \\
LM-Nav (without attack) & EnvLarge-10 & 0.8 \\
LM-Nav (with attack) & EnvLarge-10 & 0.3\scriptsize{(\textcolor{red}{\textdownarrow 62.50\%})} \\
\bottomrule
\end{tabular}
}
\label{tab:LMnav}
\end{table}

\bibliographystyle{plainnat}
\bibliography{references}